%% file: main.tex
\documentclass[runningheads]{llncs}

\usepackage{eccv}

\usepackage{eccvabbrv}

\usepackage{graphicx}
\usepackage{booktabs}
\usepackage{multirow}
\usepackage{xcolor}
\usepackage{soul}
\usepackage{algorithm}
\usepackage{algpseudocode}
\usepackage{amsmath}
\usepackage{extarrows}

\usepackage[accsupp]{axessibility}  %

\usepackage{hyperref}

\usepackage{orcidlink}

\newcommand*\ruleline[2]{\par\noindent\raisebox{.8ex}{\makebox[{#1}]{\hrulefill\hspace{1ex}\raisebox{-.8ex}{#2}\hspace{1ex}\hrulefill}}}

\begin{document}

\title{TurboEdit: Instant text-based image editing}

\author{Zongze Wu\orcidlink{0000-0001-9190-1717} \hspace{2mm}
Nicholas Kolkin\orcidlink{0000-0003-1233-1969} \hspace{2mm}
Jonathan Brandt\orcidlink{0009-0007-4769-0343} \hspace{2mm} \\
Richard Zhang \orcidlink{0000-0003-2507-4674} \hspace{2mm}
Eli Shechtman\orcidlink{0000-0002-6783-1795} 
}

\authorrunning{Z.~Wu et al.}

\institute{ Adobe Research\\
\email{\{zongzew, kolkin, jbrandt, rizhang, elishe\}@adobe.com}}

\maketitle
\input{0_abstract}
\input{fig/teaser}
\input{1_introduction}

\input{2_related}

\input{3_method}

\input{4_experiments}

\input{5_limitations}

\input{6_conclusion}

\bibliographystyle{splncs04}
\bibliography{egbib}
\clearpage

\input{7_sup}

\end{document}

%% file: 0_abstract.tex
\begin{abstract}

We address the challenges of precise image inversion and disentangled image editing in the context of few-step diffusion models. We introduce an encoder based iterative inversion technique. The inversion network is conditioned on the input image and the reconstructed image from the previous step, allowing for correction of the next reconstruction towards the input image.
We demonstrate that disentangled controls can be easily achieved in the few-step diffusion model by conditioning on an (automatically generated) detailed text prompt. To manipulate the inverted image, we freeze the noise maps and modify one attribute in the text prompt (either manually or via instruction based editing driven by an LLM), resulting in the generation of a new image similar to the input image with only one attribute changed. It can further control the editing strength and accept instructive text prompt. Our approach facilitates realistic text-guided image edits in real-time, requiring only 8 number of functional evaluations (NFEs) in inversion (one-time cost) and 4 NFEs per edit. Our method is not only fast, but also significantly outperforms state-of-the-art multi-step diffusion editing techniques.

  \keywords{Diffusion Models \and Text-Guided Image Editing}
  \vspace{-8mm}
\end{abstract}

%% file: fig/teaser.tex
\begin{figure}[!htb]
	\centering
	\setlength{\tabcolsep}{1pt}	
	\begin{tabular}{cccccc}

		   & {\tiny Man$\rightarrow$Fox} & & {\tiny Dog$\rightarrow$Chihuahua} & & {\tiny Flower$\rightarrow$Sunflower}  \\
		\includegraphics[width=0.16\columnwidth]{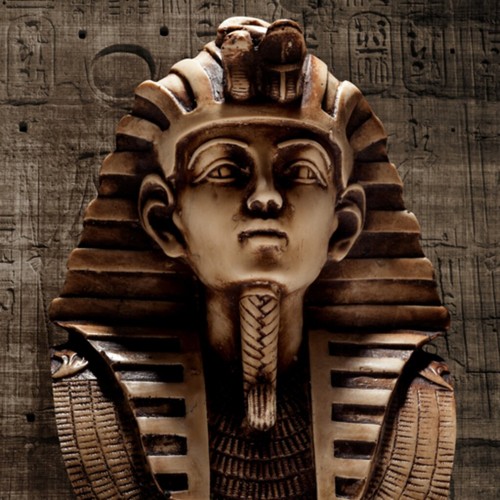} &
            \includegraphics[width=0.16\columnwidth]{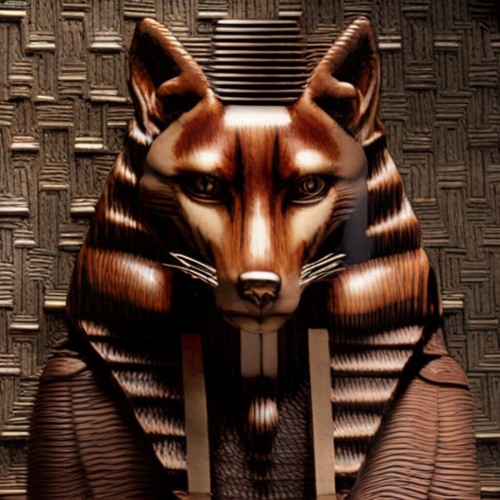} &
		\includegraphics[width=0.16\columnwidth]{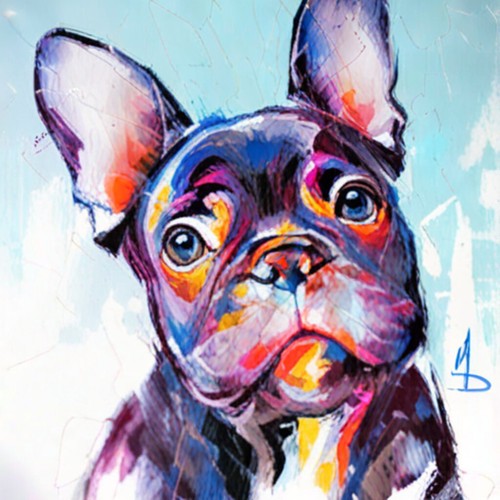} &
            \includegraphics[width=0.16\columnwidth]{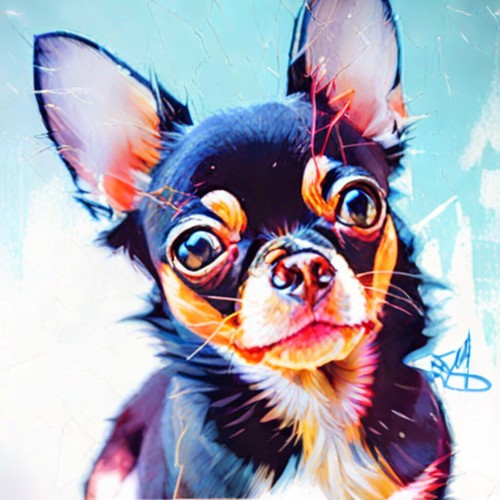} &
		\includegraphics[width=0.16\columnwidth]{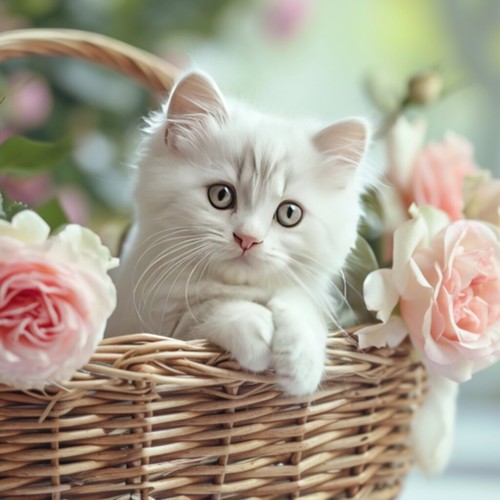} &
            \includegraphics[width=0.16\columnwidth]{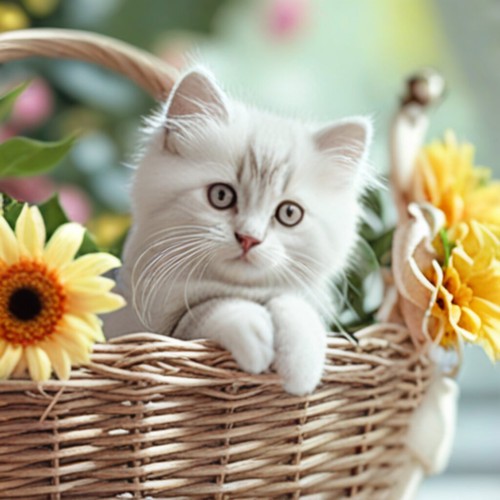} 
		\\
		   & {\tiny + Scarf } & {\tiny + Leather Jacket} & {\tiny Short Hair (0.5) } & {\tiny Short Hair (1) } & {\tiny Short Hair (1.5) }  \\
		\includegraphics[width=0.16\columnwidth]{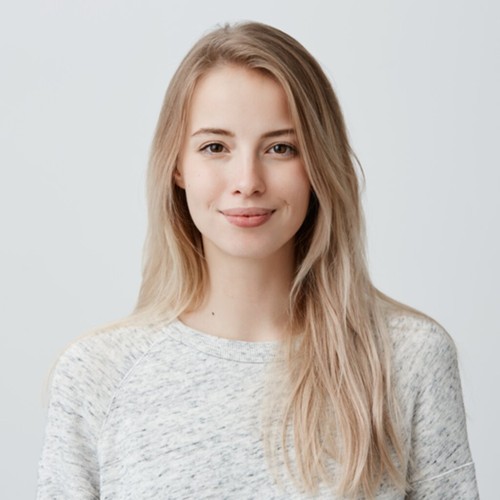} &
            \includegraphics[width=0.16\columnwidth]{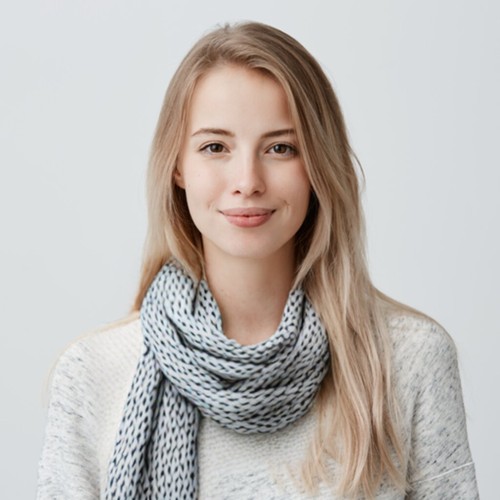} &
		\includegraphics[width=0.16\columnwidth]{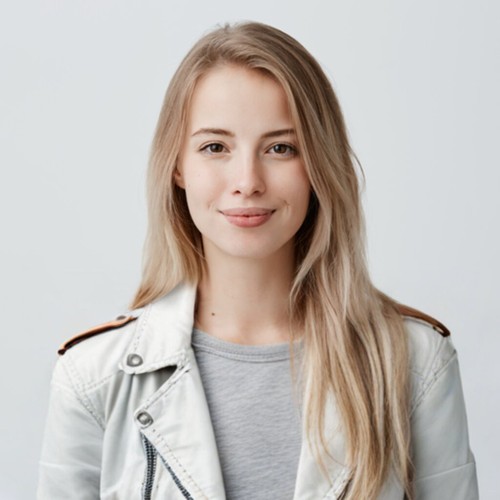} &
            \includegraphics[width=0.16\columnwidth]{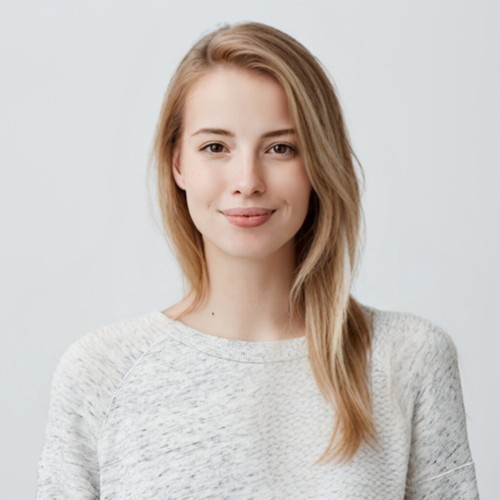} &
		\includegraphics[width=0.16\columnwidth]{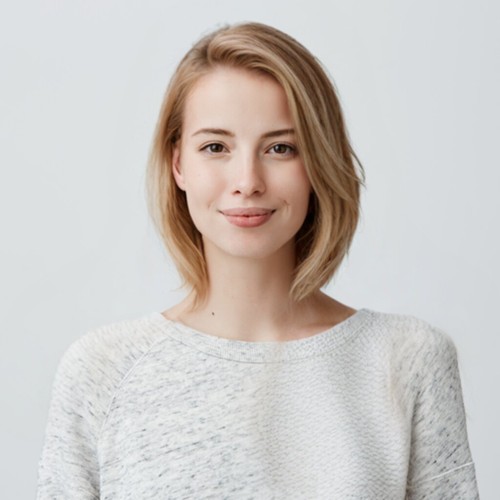} &
            \includegraphics[width=0.16\columnwidth]{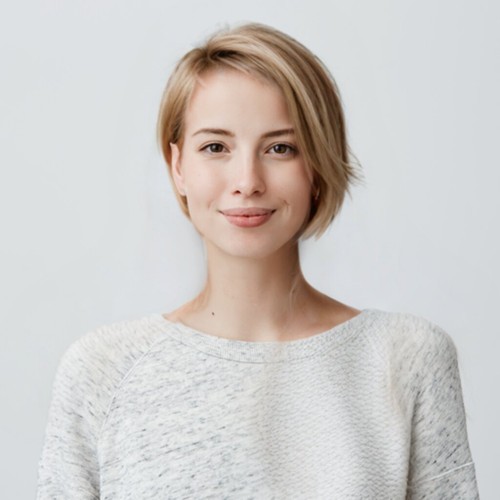} 
		\\
   \multicolumn{2}{c}{\tiny Replace his hair color to blue } &\multicolumn{2}{c}{\tiny Change the kitchen to desert} & \multicolumn{2}{c}{\tiny Make the image to child’s drawing}   \\
		\includegraphics[width=0.16\columnwidth]{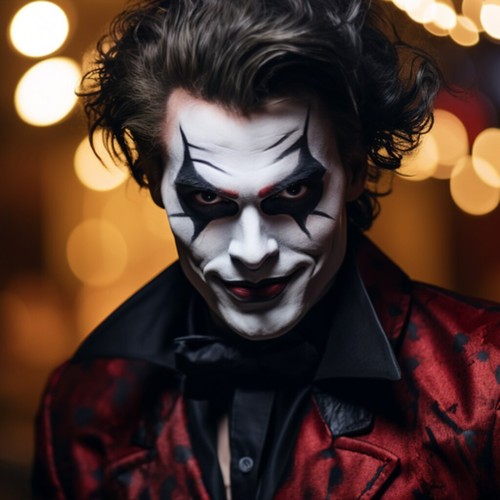} &
            \includegraphics[width=0.16\columnwidth]{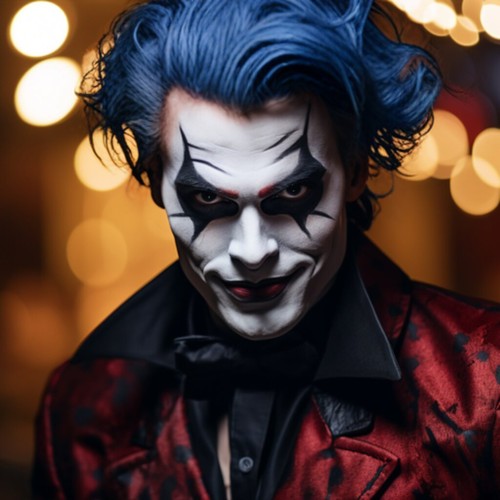} &
		\includegraphics[width=0.16\columnwidth]{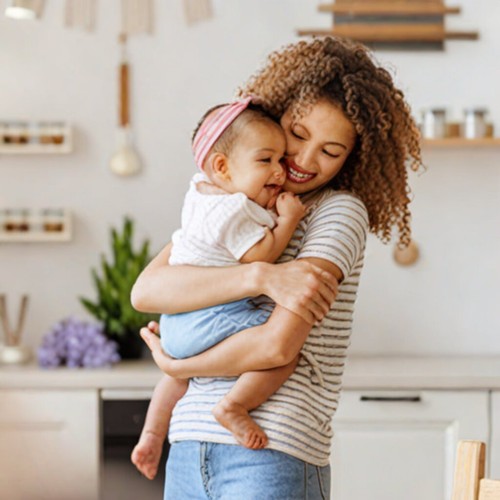} &
            \includegraphics[width=0.16\columnwidth]{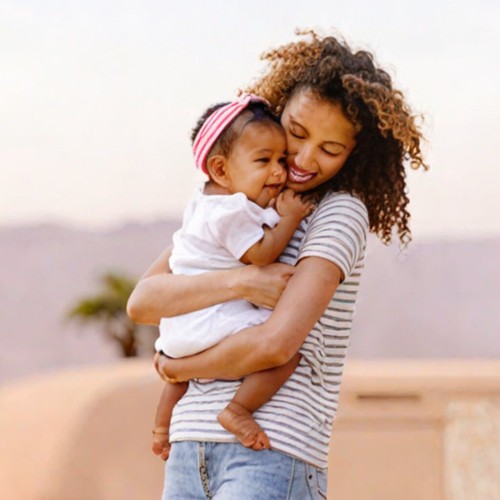} &
		\includegraphics[width=0.16\columnwidth]{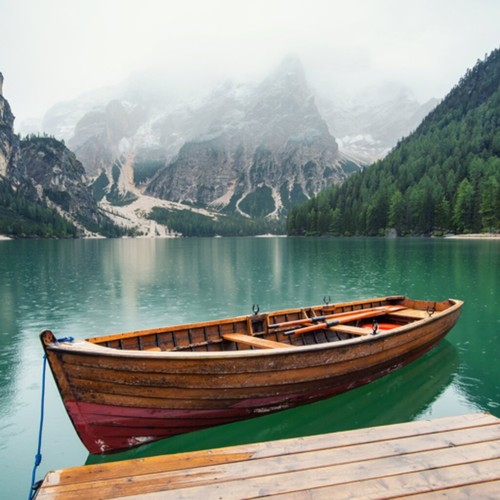} &
            \includegraphics[width=0.16\columnwidth]{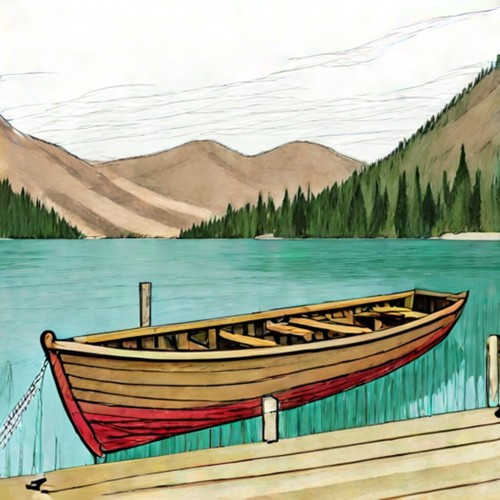} 
		\\
  
	\end{tabular}
	\caption{We present a novel real-time text-based disentangled real image editing method built upon 4-step SDXL Turbo. Our method can handle both realistic and artistic images, supports manual or instruction-based prompt manipulation, and allow users to control the editing strength. We further show multi-attribute editing and continuous editing in Supplementary Fig~\ref{fig:multi_attribute}.
    }
	\label{fig:teaser}
 \vspace{-5mm}
\end{figure}

%% file: 1_introduction.tex
\section{Introduction}
\label{sec:intro}

Large text-to-image diffusion models \cite{rombach2022high,ramesh2022hierarchical,podell2023sdxl,saharia2022photorealistic,pernias2023wurstchen} have demonstrated remarkable ability to generate photorealistic and artistic images based on a text prompt, which allows people to visually express their ideas via natural language. Many methods try to repurpose the text-to-image diffusion for the real image editing tasks \cite{huberman2023edit,wu2022unifying,mokady2023null,brack2023ledits++,ju2023direct,parmar2023zero,cao2023masactrl,xu2023inversion}. Specifically, given a real image, and a text prompt describing the target attribute, we want to disentangle alter the target attribute in the input image while keeping other attributes unchanged. 

This task can be further decomposed into two subtasks, real image inversion and disentangled image editing. Real image inversion looks for a diffusion trajectory that can precisely reconstruct the input image, which relies on DDIM inversion \cite{song2020denoising}, DDPM inversion \cite{huberman2023edit,wu2022unifying}, or their variants\cite{mokady2023null,brack2023ledits++,ju2023direct}. Disentangled image editing ensures only a single attribute change in image space, which relies on freezing attention maps\cite{hertz2022prompt,mokady2023null,parmar2023zero}, optimizing text embedding \cite{wu2023uncovering} or iterative classifier guidance \cite{kim2022diffusionclip}. Another line of works use the existing disentangled image editing methods to generate a large synthetic paired edit dataset, then train a diffusion model to accept the input image and text instruction, and output the edited images in a supervised manner, therefore they do not need image inversion or disentangle editing tricks in inference  \cite{brooks2023instructpix2pix,zhang2023hive,zhang2024magicbrush,hui2024hq}.

While these methods show promising results, their speed is bottle-necked by iterative sampling of diffusion models, which often requires $50+$ steps ($>5$ seconds) to invert a real image and $30-50$ steps ($>3$ seconds) to generate a new edit. This dramatically slows down the iterative and trial-and-error nature of image editing.

Fortunately great strides have recently been made in the methods to distill diffusion models into more efficient variants which generate images in just 1-4 steps, taking less than half a second \cite{luo2023latent,sauer2023adversarial,yin2023one,lin2024sdxl}. Few-step diffusion models pose new challenges to real image editing, as methods developed for multi-step diffusion models cannot be effectively applied. \textbf{The primary goal of our work is addressing these challenges and enabling high quality image editing using fast few-step diffusion models.}

In particular, DDIM inversion \cite{song2020denoising}, commonly used when inverting real images into the noise space of diffusion models, requires small step sizes and multiple steps ($>50$). Using only 4 steps of DDIM inversion results in blurry reconstructed images and a significant loss of detailed structure. The inverted noise containing excessive information about the input image, which deviates significantly from Gaussian noise, the true distribution of the diffusion model's noise space. This distribution shift limits the ability for large structural changes and creates pronounced artifacts in Supplementary Fig~\ref{fig:ddim}. 

Furthermore, while attention-based methods \cite{hertz2022prompt,mokady2023null,parmar2023zero} are widely used in diffusion image editing, their direct application to few-step diffusion models poses challenges. These methods often involve freezing self-attention and cross-attention maps to preserve structural similarity between the source and target images. Typically, attention maps are only frozen in the early generation steps to facilitate large structure changes and prevent artifacts. However, in the context of the one-shot diffusion model, attention control methods exert an overly restrictive influence on the generation process, leading to insufficient changes in the image space (horse to unicorn) or the introduction of artifacts (fox to dog). Even when applied solely during the initial generation steps of a four-step diffusion model, attention control can lead to either inadequate preservation of structure or the occurrence of artifacts, particularly in cases where the editing necessitates significant structural alterations in Supplementary Fig~\ref{fig:attention}.

TurboEdit confronts both challenges, offering not only a fast feed-forward mechanism for accurately inverting real images, but also a simple and efficient mechanism to make disentangled edits. Our first contribution is an inversion network that predicts noise to reconstruct the input image, and is trained to iteratively correct the reconstruction image condition on the reconstruction in previous step. We demonstrate the effectiveness of our method on complex scenes with 2-3 objects. To the best of our knowledge, this is the first encoder based diffusion model inversion method.

Our second contribution is an analysis of an emergent property of the diffusion distillation process. Namely, we show that distillation leads to disentangled adherence to long detailed text prompts, allowing for precise attribute manipulation. By changing one attribute in the long detailed text prompt, only the corresponding attribute in the image space is altered, enabling users to easily edit the text prompt and obtain the desired disentangled edit image. This technique is very simple and requires no additional implementation effort, making it highly practical. We further show that the strength of editing could be controlled by a linear interpolation of the detailed text embedding. To allow users to input instructive text prompt, we utilize large language model to transform the descriptive source prompt and instructive prompt to descriptive target prompt, then input the target prompt to diffusion model, which enable instructive control in text-to-image diffusion model. Combined, these elements enable TurboEdit to make image edits that are high-quality, real-time, text-based, and disentangled. 

Our method only requires 8 number of functional evaluations (NFEs) in inversion (one-time cost) and 4 NFEs per edit, compared to 50 NFEs in inversion and 30-50 NFEs per edit for methods based on multi-steps diffusion models. Despite being significantly faster ($<0.5$ seconds instead of $>3$ seconds per edit), our method shows better text image alignment and background preservation compared to methods based on multi-step diffusion models in both descriptive and instructive text prompt setting.

%% file: 2_related.tex
\section{Related Works}
\subsection{Text-To-Image diffusion models}

Large text-to-image diffusion models \cite{rombach2022high,ramesh2022hierarchical,podell2023sdxl,saharia2022photorealistic,pernias2023wurstchen} transform random Gaussian noise into natural images conditioned on text prompts through iterative denoising. While they produce high-quality images, they require 30-50 denoising steps and over 3 seconds per generation. Recently, distillation methods have been developed to speed up the process, enabling image generation in just 1-4 steps and less than 1 second \cite{luo2023latent,sauer2023adversarial,yin2023one,lin2024sdxl}. These few-step distilled models offer the best balance between speed and quality, making them ideal for real-time image editing.

\subsection{Text-Based Image Editing}

To edit an existing image using a text-to-image diffusion model, we first need to map a real image onto the reverse diffusion trajectory, ensuring the inverted images resemble the input image. Current methods often use DDIM inversion \cite{song2020denoising}, requiring over 50 small diffusion steps. Null-Text inversion \cite{mokady2023null} improves DDIM's reconstruction quality by optimizing the null text token at each timestep, but this process is time-consuming. DDPM inversion \cite{huberman2023edit,wu2022unifying} uses an over-complete noise representation for an inversion procedure to prevent error accumulation, resulting in perfect reconstruction for the input text prompt. LEdits++ \cite{brack2023ledits++} extends this idea with a higher-order solver and an automatic method to mask edits, preserving most of the image's identity.

\subsection{Disentangled Editing with Generative Models}

Attention-based methods \cite{hertz2022prompt,parmar2023zero} freeze self-attention and cross-attention maps to preserve structural similarity between the source and target images. Typically, attention maps are only frozen in the early generation steps to facilitate large structure changes and prevent artifacts. Similarly Cao \textit{et al.} \cite{cao2023masactrl}, propose a mechanism to directly share attention maps with the input image during generation, but this makes editing the texture of the foreground challenging and makes this method more appropriate to altering pose.

Other methods optimize text embedding tokens\cite{mokady2023null,wu2023uncovering}, or the model itself \cite{kawar2023imagic}; however the expensive optimization process per input image runs counter to our goal of real-time image editing. Another approach, pioneered by InstructPix2Pix, is to finetune a model on synthetic instruction-based image editing datasets\cite{brooks2023instructpix2pix}. While efficient, this method has poor disentanglement and requires expensive dataset preparation and training. In contrast our trained inversion network is easy to train, requires no training data beyond the original diffusion model, and offers more disentangled, identity preserving, edits.

%% file: 3_method.tex
\section{Method}
\subsection{Preliminary}

The forward diffusion process gradually turns a clean image $x_0$ into white Gaussian noise $x_T$ by  adding Gaussian noise $\epsilon_t$ to it,

\begin{equation} \label{equation:alpha}
x_t=\sqrt{\bar{\alpha}_t}x_0+\sqrt{1-\bar{\alpha}_t}\epsilon_t
\end{equation}

where $\bar{\alpha}_t$ controls noise schedule and $\epsilon_t$ is Gaussian noise. A network $\hat{\epsilon}_\theta$ is trained to predict $\epsilon_t$ given $x_t$, text prompt $c$ and time step $t$ with objective

\begin{equation} \label{equation:sampling}
L(\hat{\epsilon}_\theta)=\mathbb{E}_{x_0 \sim q;\epsilon_t \sim \mathcal{N}(0,1)}[\parallel \hat{\epsilon}_\theta(x_t, c, t) - \epsilon_t \parallel^2]
\end{equation}

We can easily rewrite the formulation from noise prediction to sample prediction 

\begin{equation} \label{equation:zero}
x_{0,t}=\frac{x_t-\sqrt{1-\bar{\alpha}_{t}}\hat{\epsilon}_\theta(x_t,c, t)}{\sqrt{\bar{\alpha}_{t}}}
\end{equation}

It usually takes 20-50 steps from a sampled Gaussian noise $x_T$ to a clean image $x_0$. With the developed of distillation methods \cite{luo2023latent,sauer2023adversarial,yin2023one,lin2024sdxl}, few-step diffusion models can obtains high quality images in 1-4 steps.

\subsection{Single Step Image Inversion}

Current diffusion-based methods for real image editing have shown promising results in achieving high-quality disentangled edits \cite{cao2023masactrl,tsaban2023ledits,brack2023ledits++,brooks2023instructpix2pix,parmar2023zero,mokady2023null}. However, these methods, which rely on multi-step diffusion models \cite{rombach2022high,podell2023sdxl}, are hindered by their computational demands, with each edit requiring at least 4-5 seconds, making them unsuitable for interactive applications. Moreover, these methods cannot be directly applied to few-step diffusion models due to fundamental differences in their design. For instance, many diffusion-based editing approaches rely on the DDIM inversion \cite{song2020denoising} or DDPM inversion \cite{huberman2023edit} \ to project real images into diffusion noise space. However, DDIM inversion's requirement for a small step size and a large number of inversion steps is inherently at odds with the design principles of few-step diffusion models. While DDPM inversion overfits to the input image and produce significant amount of artifact in the edited image. As illustrated in Supplementary Fig~\ref{fig:ddim}, both DDIM and DDPM inversion yield suboptimal editing results when applied to few-step inversion steps. 

Several works \cite{yin2023one,sauer2023adversarial} utilize adversarial loss to distill a multi-step diffusion model, and make the optimization target of few-shot diffusion models similar to GANs \cite{goodfellow2014generative}. This inspire us to draw on ideas from the GAN inversion literature, where encoder based methods have been shown to be efficient and reliable \cite{richardson2021encoding,tov2021designing,alaluf2021restyle}.

Consider a generator $G$ (in our case SDXL-Turbo \cite{sauer2023adversarial}) which accepts time step $t$, text prompt $c$, and noisy image $x_t=x_0+\epsilon_t$ and outputs a reconstructed image $x_{0,t}$. Given this, we predict the clean image $x_{0,t}$ from a noisy version as $x_{0,t}=G(x_t,c,t)$. We begin designing our inversion network using a single step approach where $t=T$. We train an inversion network $F_\text{single}$ to predict $x_T$, such that when inputting  $x_T$ to $G$,  $x_{0,t}$ will match $x_0$. Leading to the loss function:

\begin{equation}
\begin{split} 
L(F_\text{single})&=\mathbb{E}_{x_0 \sim q}[\parallel x_0-G(F_\text{single}(x_0,c,T), c, T) \parallel^2]
\end{split} 
\end{equation}

The inversion network $F_\text{single}$ is initialized from $G$ (SDXL-Turbo), and the generator $G$ is frozen during training. The information of input image $x_0$ is stored in text prompt $c$ (global information) and initial noise $x_T=F_\text{single}(x_0,c,T)$ (spatial information). When we want to perform image editing, we use a new text prompt $c'$, and generate the edited image by

\begin{equation}
\begin{split} 
x'_{0,T}&=G(F_\text{single}(x_0,c,T), c', T) , \\
\end{split} 
\end{equation}

 Despite its simplicity, the single-step encoder method is capable of impressive semantic edits while preserving background details, surpassing the performance of DDIM and DDPM inversion methods and emerging as the sole viable option for single-step inversion in Supplementary Fig~\ref{fig:ddim}. However, its results exhibit artifacts in regions such as hands and faces. The resulting images lack sharpness and contain salt-and-pepper noise, falling short of photorealistic quality. To combat this, we extend our method to multiple inversion steps.

\input{fig/inversion}

\subsection{Multi-Step Image Inversion}

In order to enhance the quality of image reconstruction, we leverage a multi-step inversion approach which iteratively refines the reconstruction in each step, similar to the GAN inversion network proposed in ReStyle \cite{alaluf2021restyle}. The inversion network $F$ is designed to take the input image $x_0$ along with the reconstruction from the previous step $x_{0,t+1}$, and predict the injected noise $\epsilon_t$ for the current step. This injected noise $\epsilon_t$, combined with the previous reconstruction $x_{0,t+1}$, forms the new noisy image $\hat{x_{t}}$ according to equation \ref{equation:alpha}, which serves as input to $G$. Then we obtain a new reconstruction image $x_{0,t}$ according to equation \ref{equation:zero}. This yields the initial multi-step training objective: 

\begin{equation}
\begin{split} 
L(F)&=\mathbb{E}_{x_0 \sim q}[\parallel x_0-G(\hat{x_t},c,t) \parallel^2], \\
\hat{x_t}&=\sqrt{\bar{\alpha}_t}x_{0,t+1}+\sqrt{1-\bar{\alpha}_t}F(x_0,x_{0,t+1},c,t)
\end{split}
\end{equation}

It is worth emphasizing that generator $G$ takes previous reconstruction $x_{0,t+1}$ as input, therefore the loss function pushes $F$ to output an $\epsilon_t$ which improve the previous reconstruction $x_{0,t+1}$ relative to the input image $x_0$. During training, we simulate the reconstruction from the previous step $x_{0,t+1}$ using single step SDEdit \cite{meng2021sdedit}. Specifically, we add random Gaussian noise to input image to get $x_{t+1}$, then input $x_{t+1}$ to the generator to get $x_{0,t+1}$.
At maximum time step $t=T$, we use an all zeros matrix as $x_{0,t+1}$. 

Our analysis revealed that a naive implementation of this model results in predicted noise containing numerous high values ($>10$) and excessive structural information from the input image, resulting in artifacts in the reconstruction image. Moreover, changing the text prompt had minimal effect on the output image. To address these issues, we employ the reparameterization trick \cite{kingma2013auto} to constrain the injected noise to a distribution close to standard Gaussian. Instead of directly predicting the value of the injected noise, the inversion network outputs the mean and variance of each pixel, from which the injected noise is sampled. The KL loss required for this modification is:

\begin{equation} 
L_{KL}(F)=\mathbb{E}_{x_0 \sim q}[KL(F(x_0,x_{0,t+1},c,t),\mathcal{N}(0,1))]
\end{equation}

This yields the final training objective:

\begin{equation} 
L(E)=L_{MSE}(F) + \lambda * L_{KL}(F)
\end{equation}

Through experimentation, we determined that setting the $\lambda=10^{-6}$ achieved a favorable balance between reconstruction quality and editability. After training, we can perform iterative inversion as show in Fig~\ref{fig:method},  Algorithm~\ref{algorithm:inversion}. The inversion process iterates from $t=T$ to smaller $t$, with the intention of first encoding semantic information and then capturing finer details. The noise $\epsilon_t$ contain spatial information not explicitly encoded in $c$. Through experimentation, we determine that a four-step inversion is sufficient to faithfully reconstruct complex images and preserve facial identity in Figure~\ref{fig:inversion_abl}. Given $\epsilon_t$ and new text prompt $c'$, we can generate a new image resembling the input image $x_0$ while containing the target attribute in $c'$ as show in Algorithm~\ref{algorithm:edit}. In summary, the inversion process takes 8 NFE ($4\times2$) since each inversion step requires inference of both inversion network and generator. Once the image is inverted, all subsequent edit takes 4 additional NFE.

\begin{minipage}{0.46\textwidth}
\begin{algorithm}[H]
    \centering
    \caption{Iterative Inversion}
\begin{algorithmic}
\State \textbf{Input} real image $x_0$, corresponding caption $c$ 
\State  $x_{0,T+1}=0$
\State \textbf{for} t=T to 1 \textbf{do}
\State \phantom{kk} $\epsilon_t=E(t,c,x_0,x_{0,t+1})$
\State  \phantom{kk} $\hat{x_t} =\sqrt{\bar{\alpha}_t}x_{0,t+1}$
\State \phantom{kkkkk} $+\sqrt{1-\bar{\alpha}_t}F(t,c,x_0,x_{0,t+1})$
\State \phantom{kk} $x_{0,t}=G(t,c,\hat{x_t})$
\State \textbf{Return} {$\{\epsilon_T,...,\epsilon_1\}$, $\{x_{0,T},...,x_{0,1}\}$}
 \phantom{kkk}
\end{algorithmic}
\end{algorithm} \label{algorithm:inversion}
\end{minipage}
\hfill
\begin{minipage}{0.46\textwidth}
\begin{algorithm}[H]
    \centering
    \caption{Iterative Editing}
\begin{algorithmic}
\State \textbf{Input} target caption $c'$, $\{\epsilon_T,...,\epsilon_1\}$ 
\State \textbf{Optional Input} mask $m$, $\{x_{0,T},...,x_{0,1}\}$
\State  $x'_{0,T+1}=0$
\State \textbf{for} t=T to 1 \textbf{do}
\State \phantom{kk} $\hat{x_t} =\sqrt{\bar{\alpha}_t}x_{0,t+1}+\sqrt{1-\bar{\alpha}_t}\epsilon_t$
\State \phantom{kk} $x'_{0,t}=G(t,c',\hat{x_t})$
\State \phantom{kk} \textbf{if} $m$ is not None \textbf{then}
\State \phantom{kkkk} $x'_{0,t}=m*x'_{0,t}+(1-m)*x_{0,t}$

\State \textbf{Return} $x'_{0,1}$
\end{algorithmic}
\end{algorithm} \label{algorithm:edit}
\end{minipage}
\vspace{-10mm}

\subsection{Detailed Text Prompt Condition}

\input{fig/turbo}

\input{fig/interpolation}

Attention-based image editing methods \cite{hertz2022prompt,mokady2023null,parmar2023zero} preserve the structural similarities between source and target images by freezing self-attention and cross-attention maps. Although they work well in regular multi-step diffusion models, we show that it over constrains the structure of target images and trend to produce artifacts in both single-step or four-step diffusion model in Supplementary Fig~\ref{fig:attention}.

To enable text guided image editing in few-step diffusion models, we propose an extremely simple method. Our intuition is that if the text prompt is highly detailed and encompasses semantic information across various aspects, modifying a single attribute in the text prompt will result in only a minor change in the text embedding. Consequently, the source and target sampling trajectories remain sufficiently close, resulting in generated images that are nearly identical except for the modified attribute in Fig~\ref{fig:turbo}. The same intuition applies to real image edit as we show in Supplementary Fig~\ref{fig:long_short_prompt}. Moreover, we can linear interpolate the detailed source and target text embeddings and generate a smooth interpolation in image space in Fig~\ref{fig:teaser} and \ref{fig:interpolation}. Although it is hard for users to write a long text prompt, we can easily utilize ChatGPT \cite{ouyang2022training} to expand a short text prompt (e.g., "please describe an image of a \{short user-provided caption\} in detail"), or use LLaVA \cite{liu2024visual} generate detailed captions of a given image. 

A concurrent work also show image editing capability sololy based on text embedding \cite{yu2024uncovering} without freezeing attention map. We want to highlight the difference between these two methods. To perform object replacement or style control, they substitute the keyword embeddings in the text embedding space, while we use long detailed text prompts and substitute the keyword directly in the text space. To control the editing strength, they rescale the weight of the descriptive word embedding or use singular value decomposition to discover editing direction in text embedding space, while we directly linearly interpolate the source and target text embedding.

\subsection{Local Mask}

To facilitate localized edits, our method permits users to upload a binary mask indicating the region to be edited. We first Gaussian Blur the mask, then resize it to match the latent image size ($64\times64$). Subsequently, we retain the edited image $x'_{0,t}$ at time step $t$ only within the masked region, employing the inverted image $x_{0,t}$ for the remainder of the image, as outlined in Algorithm ~\ref{algorithm:edit}.

To provide an initialization of the masks, we propose utilizing a rough attention mask to denote the edited region. Drawing inspiration from the local blend mode in prompt2prompt\cite{hertz2022prompt}, we automatically extract the attention mask at resolution $16\times16$ for words that only exist in either source or target prompt, sum it over channel dimension, divide it by its maximum value. This process yields a single-channel attention mask with values ranging from 0 to 1, where the edited region is characterized by high attention values and the unchanged region by low attention values. By default, we set the threshold at 0.6 and convert the continuous attention mask to a binary mask. Users can interactively adjust the threshold to control the size of the mask, as each edit (4 steps) requires less than 0.5 seconds. Although the attention mask is very rough, we show it can significantly improve the background and identity preservation in Supplementary Fig~\ref{fig:mask}. In our figures we only use rough attention mask rather than accurate manual mask. 

It is important to clarify that our approach utilizes attention masks solely to constrain the editing region, which is different from freezing attention maps for structural alignment in prompt2prompt\cite{hertz2022prompt}. Our method is orthogonal to attention freezing and can be combined with it. However, by default, we do not freeze attention maps at any time step, as doing so strongly constrains the object structure and tends to introduce artifacts in few-step diffusion models in Supplementary Figure~\ref{fig:attention}.

\subsection{Instruction-based Editing} \label{sec:instruct}

In many editing scenarios, users need to change multiple words from the source prompt to get the desired target prompt. For instance, when users want to change an image of a little dog to an image of a little cat, they need to change the word ``dog'' to ``cat'' and ``puppy'' to ``kitten'', which can be cumbersome and unappealing.

Fortunately, instruction tuning and semantic editing in text space is well explored for large language models (LLM). We start with a base instruction like ``please make the smallest change possible to the following sentence, but...'' and then users only need to add task specific instruction like ``change the dog to cat.''  We concatenate the base instruction, user instruction, and source prompt, and input them to an LLM. The LLM figures out the best way to make the edit and generate the target prompt. In this way, the complex text edit is handled by the LLM, and users only need to input simple short instructions. To make this process more efficient and save memory, we reuse LLaVA as our LLM, but any instruction-tuned LLM could be swapped in. LLaVA is built on top of a LLM Vicuna \cite{vicuna2023} and can still do text editing tasks even after being fine-tuned for vision and language tasks. Experiments show this simple method works well in Fig~\ref{fig:teaser}, \ref{fig:instruct} and Supplementary Table~\ref{tab:instruct_comparison}.

%% file: fig/inversion.tex
\begin{figure}[t]
\centering
\includegraphics[width=1.\columnwidth]{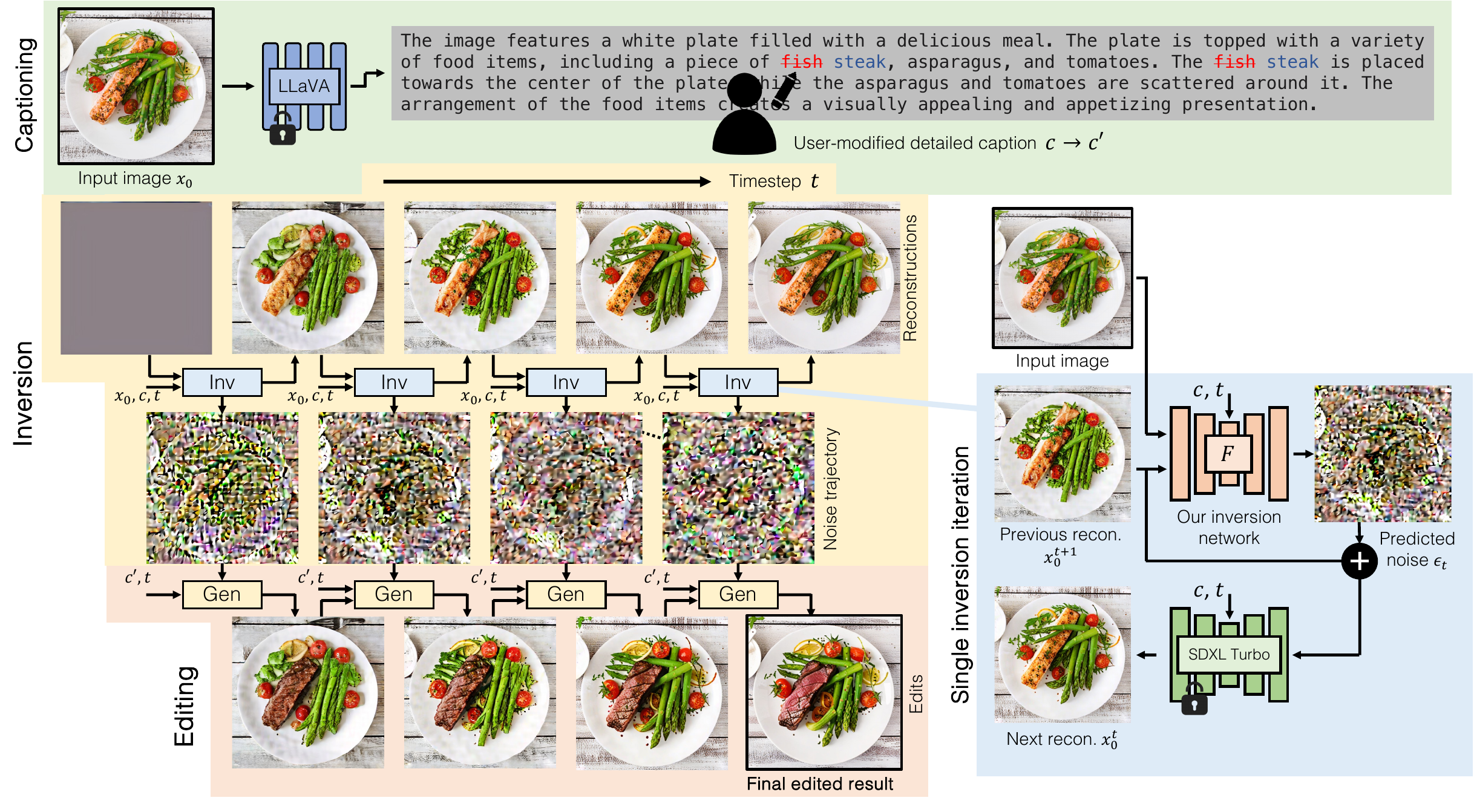}
\caption{\small Given an input real image $x_0$, we utilize the LLaVA to generate a detailed caption $c$. Users can modify $c$ to create a new text prompt $c'$. The inversion process begins by feeding the $x_0$, $c$, current time step $t$, and a previously reconstructed image $x_{0,t+1}$ (initialized as a zero matrix) into the inversion network. This network then predicts the noise $\epsilon_t$, which is subsequently input into a frozen SDXL-Turbo model to generate the new reconstruction image $x_{0,t}$. Given the final inverted noise $\epsilon_t$, along with $c$, we can use SDXL-Turbo to create an inversion trajectory and reconstruct $x_{0,0}$, which is very similar to $x_0$. Using the same noises $\epsilon_t$ and slightly different text prompt $c'$, starting from $t=T$ to smaller $t$, the editing trajectory will be very similar to the inversion trajectory, and the generated image will closely resemble the input image, differing only in the specified attribute in $c'$.}
\label{fig:method}
\vspace{-0.5cm}
\end{figure}

%% file: fig/turbo.tex
\begin{figure}[tb]
	
	\setlength{\tabcolsep}{1pt}	
	\begin{tabular}{ccccccc}
		 & {\scriptsize } & {\scriptsize \textcolor{brown}{Black $\rightarrow$ Blond}} & {\scriptsize \textcolor{green}{Trench $\rightarrow$ Down}} &{\scriptsize \textcolor{purple}{+ Hat}} &{\scriptsize \textcolor{blue}{+ Snow}} \\
		\rotatebox{90}{\scriptsize \phantom{kkk} Short Text} &
		\includegraphics[width=0.18\columnwidth]{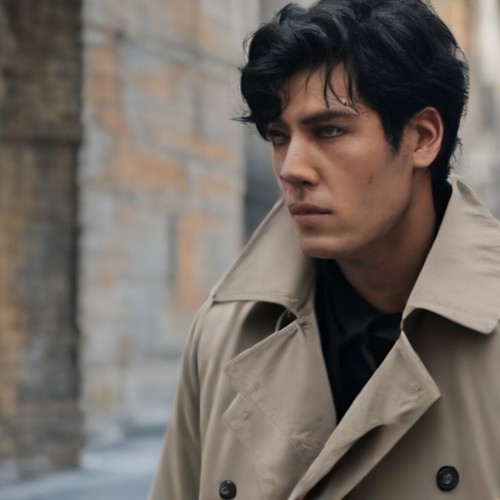} &
		\includegraphics[width=0.18\columnwidth]{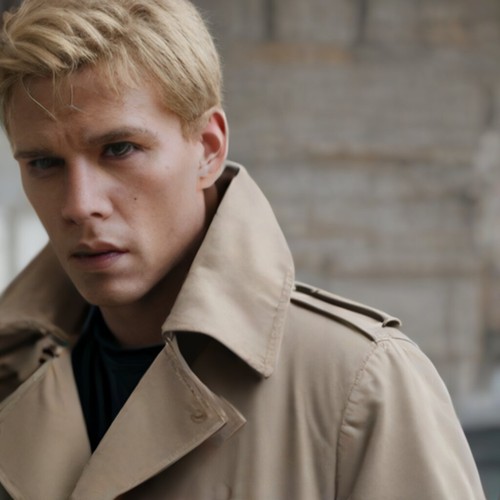} &
		\includegraphics[width=0.18\columnwidth]{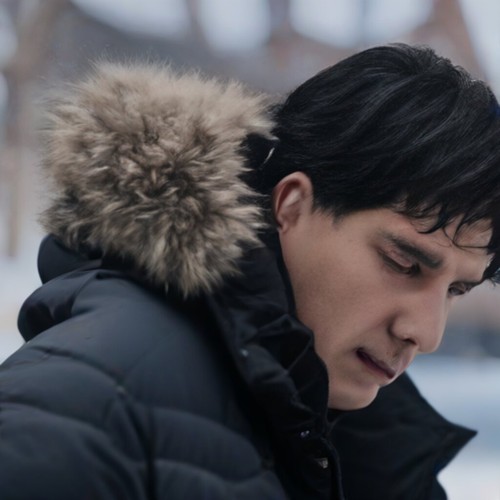} &
		\includegraphics[width=0.18\columnwidth]{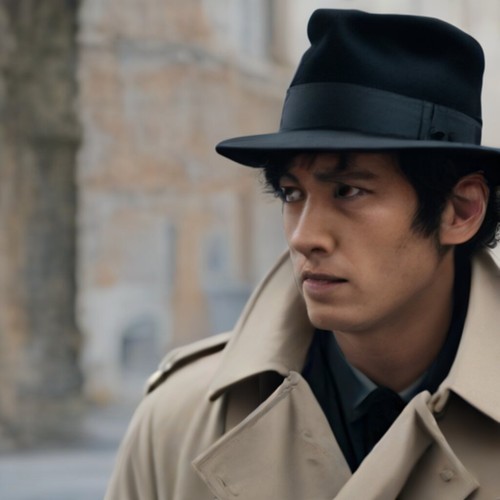} &
		\includegraphics[width=0.18\columnwidth]{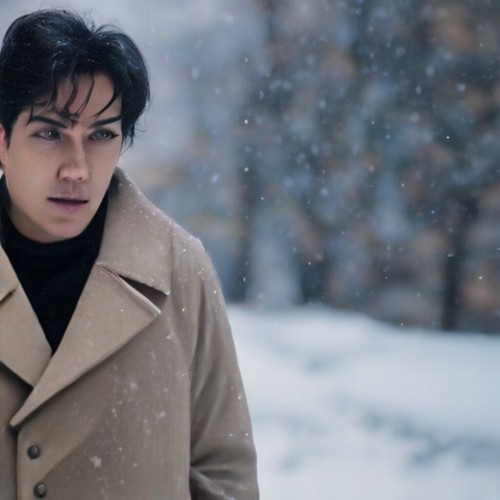} 
		\\

            \rotatebox{90}{\scriptsize \phantom{kk} Detailed Text} &
		\includegraphics[width=0.18\columnwidth]{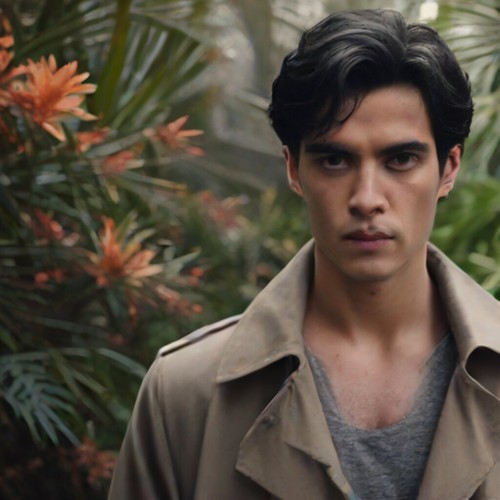} &
		\includegraphics[width=0.18\columnwidth]{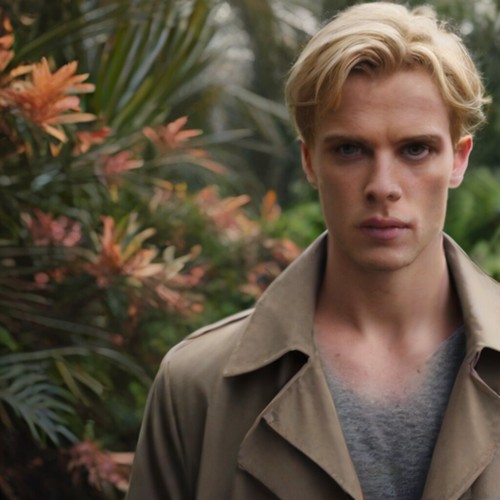} &
		\includegraphics[width=0.18\columnwidth]{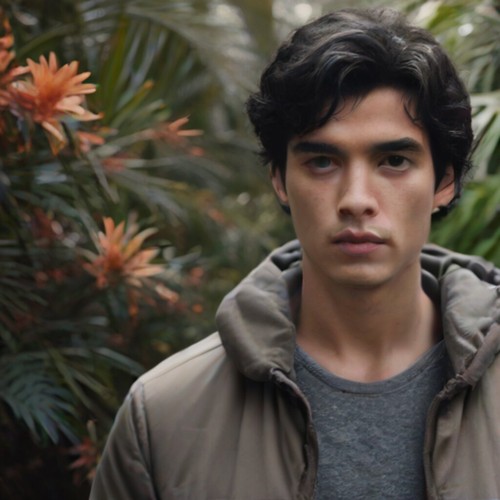} &
		\includegraphics[width=0.18\columnwidth]{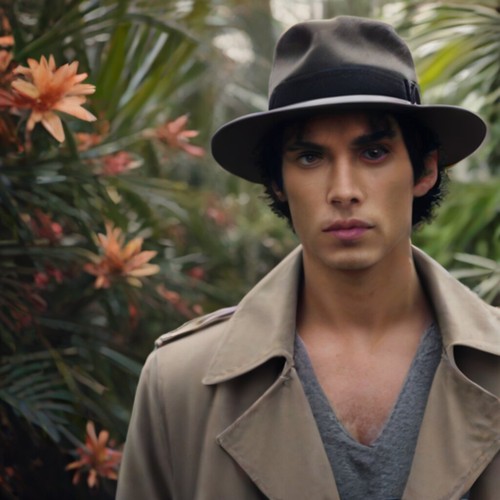} &
		\includegraphics[width=0.18\columnwidth]{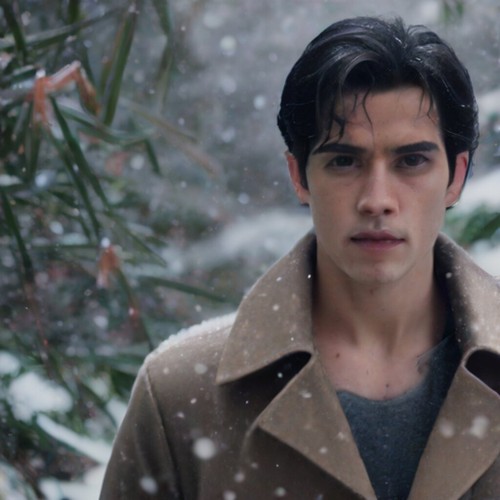} 
		\\

	\end{tabular}
 \vspace{-2mm}
	\caption[Turbo]{ When presented with a concise source text prompt, minor edits in the text space can lead to substantial layout and structural changes in the image space. Conversely, making small text edits in a detailed text prompt tends to cause more disentangled changes in the image space. The results are from single step image generation with the same random seed. The captions and color-coded modification areas are provided below.}

    \scriptsize{\textcolor{red}{Short Text:} ``a man wearing a \textcolor{green}{trench coat} \textcolor{purple}{(, a hat,)} with \textcolor{brown}{black hair} \textcolor{blue}{(in heavy snow)}.''}\\
   \scriptsize{\textcolor{red}{Detailed Text:} ``a young man wearing a brown \textcolor{green}{trench coat} \textcolor{purple}{(and a hat,)} and grey t-shirt with \textcolor{brown}{black hair}, standing in front of subtropical flowers \textcolor{blue}{(in heavy snow)}. He is looking directly at the camera, giving a sense of focus and determination. The coat is open, revealing the man's attire underneath. The overall scene is well-lit, with the man being the main subject of the image.''}
	\label{fig:turbo}
  \vspace{-5mm}
\end{figure}

%% file: fig/interpolation.tex
\begin{figure}[tb]
	\centering
	\setlength{\tabcolsep}{1pt}	
	\begin{tabular}{ccccccc}

		  {\scriptsize Luxurious Sofa} & \multicolumn{3}{c}{$\xlongleftrightarrow{\hspace*{6cm}}$} & {\scriptsize Wood Chair} \\
		\includegraphics[width=0.18\columnwidth]{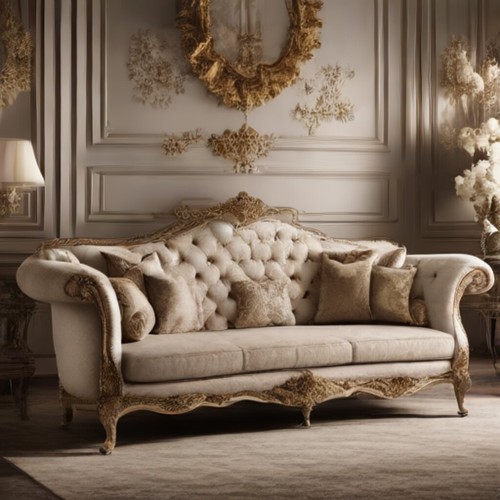} &
		\includegraphics[width=0.18\columnwidth]{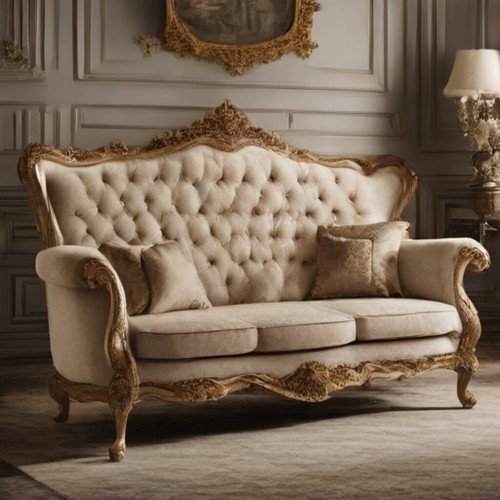} &
		\includegraphics[width=0.18\columnwidth]{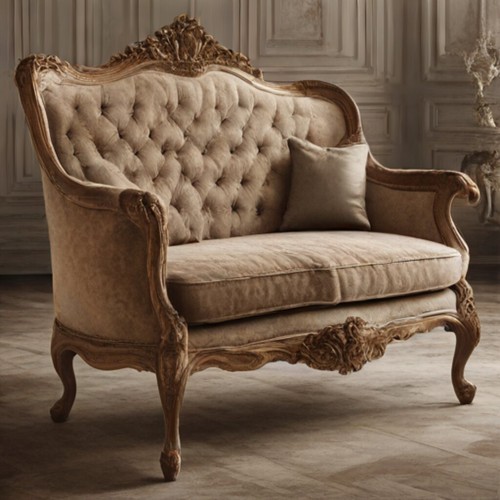} &
		\includegraphics[width=0.18\columnwidth]{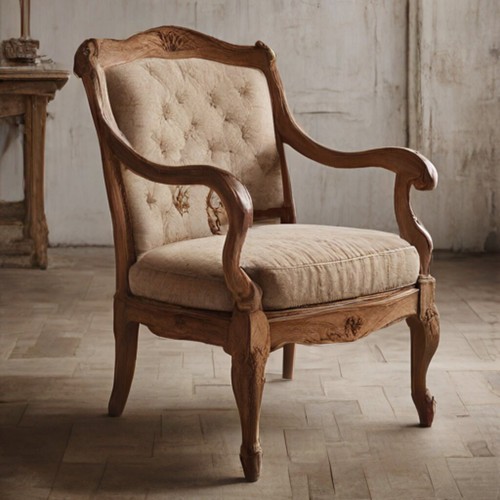} &
		\includegraphics[width=0.18\columnwidth]{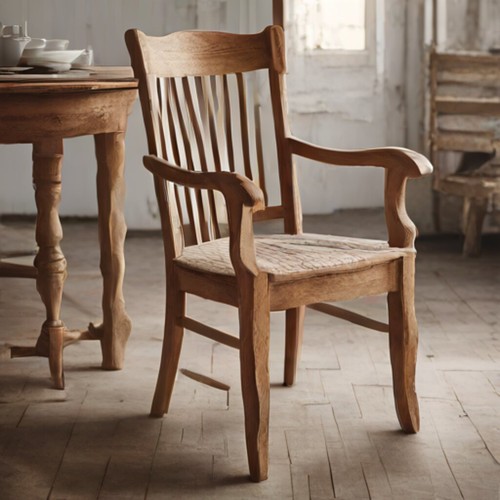} 
		\\
  
		  {\scriptsize SUV} & \multicolumn{3}{c}{$\xlongleftrightarrow{\hspace*{6cm}}$} &{\scriptsize Bike} \\
		\includegraphics[width=0.18\columnwidth]{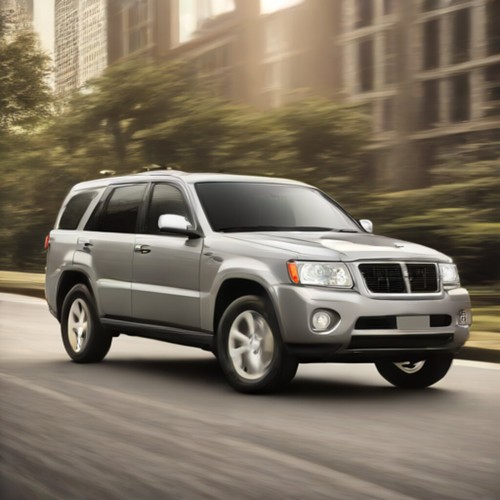} &
		\includegraphics[width=0.18\columnwidth]{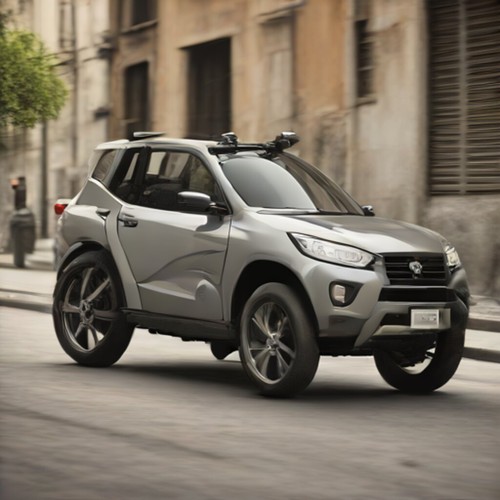} &
		\includegraphics[width=0.18\columnwidth]{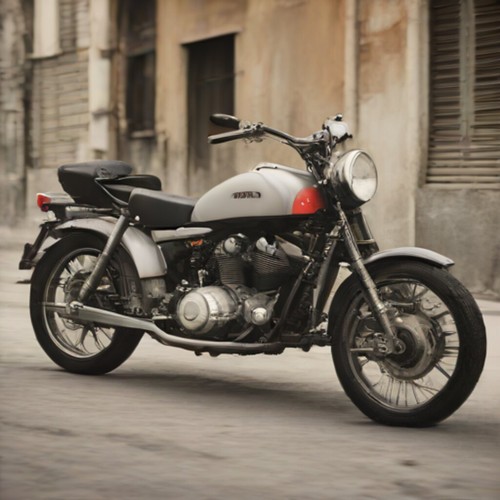} &
		\includegraphics[width=0.18\columnwidth]{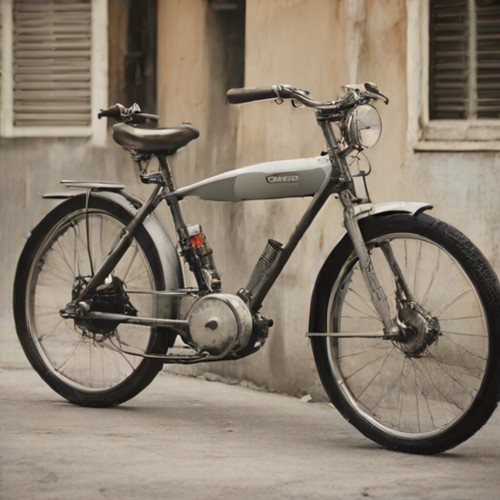} &
		\includegraphics[width=0.18\columnwidth]{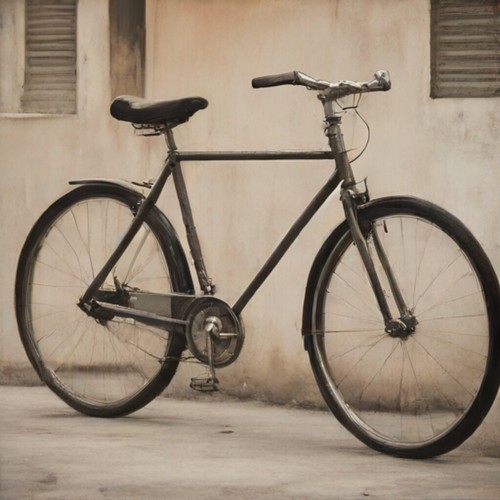} 
		\\
	\end{tabular}
	\caption[Turbo]{Given a detailed source text and corresponding target text, we can interpolating the text embeddings and generate a smooth interpolation in image space even for large structure change. }
	\label{fig:interpolation}
  \vspace{-5mm}
\end{figure}

%% file: 4_experiments.tex
\section{Experiments}

\subsection{Training Details}

In order to address computational and storage constraints, we select only 250k images larger than 512$\times$512 pixels from an internal dataset, performing center cropping to obtain square images and resizing them to 512$\times$512 pixels. To generate detailed captions, we input these images into the LLaVA model with the prompt ``please describe the image as detailed as possible, including layout, objects, and color''. Subsequently, we precompute the image and text embeddings for the SDXL-Turbo model before training. The inversion network $F$ is initialized from the SDXL-Turbo model, while the generator $G$ (also SDXL-Turbo) remains fixed throughout training. Training is conducted over four different time steps (1000, 750, 500, 250), consistent with the approach employed in SDXL-Turbo \cite{sauer2023adversarial}. We utilize a learning rate of  $10^{-5}$ and a batch size of 10, achieving model convergence within a day using eight A100 GPUs.

\begin{table*}
  \centering
  \resizebox{\textwidth}{!}{
  \begin{tabular}{lrrrrrrrrrrrr} 
    \toprule
    Method &
      \multicolumn{4}{c}{Background Preservation} & &
      \multicolumn{2}{c}{CLIP Similarity} & &
      \multicolumn{3}{c}{Efficiency}\\
        \cline{2-5}  \cline{7-8} \cline{10-12} \vspace{-2mm} \\  
       &  PSNR  & \phantom{k} LPIPS & \phantom{k}MSE & \phantom{k}SSIM &&  Whole   &  Edited   && Inverse  & Forward  & Steps  \\
&  $\uparrow$\phantom{kk}  & ${10^3}\downarrow$ &  $10^4\downarrow$ & $10^2\uparrow$ &&  $\uparrow$\phantom{kk}  &  $\uparrow$\phantom{kk} && (s) $\downarrow$ & (s) $\downarrow$ &  $\downarrow$\phantom{k} \\
       \cline{2-5}  \cline{7-8} \cline{10-12} \vspace{-2mm}\\
    Null-Text Inv & 27.03 & 60.67 & 35.86 & 84.11 &&  24.75 & 21.86 && 56.98 & 3.66 & 50\\
    Direct Inv & 27.22 & 54.55 & 32.86 & 84.76 && 25.02 & 22.10  && 10.14 & 4.30 & 50 \\  
    P2P-Zero & 20.44 & 172.22 & 144.12 & 74.67& & 22.80 & 20.54  && 11.33 & 12.36 & 50 \\
    MasaCtrl & 22.17 & 106.62 & 86.97 & 79.67 && 23.96 & 21.16 &&  4.14 & 4.83 & 50 \\
    Inversion-Free & 28.51 & 47.58 & 32.09 & 85.66 && 25.03 & 22.22 && \textbf{ ---} & 0.975 & 12 \\
    DDIM & 18.59 & 177.96 & 184.69 & 66.86 && 23.62 & 21.20 && \textbf{0.344} & \textbf{0.341} & \textbf{4} \\  
    Ours &  \textbf{29.52} & \textbf{44.74} & \textbf{26.08} & \textbf{91.59} && \textbf{25.05} & \textbf{22.34} && 0.668 & 0.508 & \textbf{4}\\
    \bottomrule
  \end{tabular}
  }
  \caption{Image editing comparison using descriptive text in PIE-Bench dataset. The efficiency is measured in a single H100 GPU. Our method achieves the best background preservation and clip similarity, while being significantly faster than other methods (except 4 steps DDIM).}
  \label{tab:comparison}
    \vspace{-15mm}
\end{table*}

\subsection{Quantitative Comparison}

The PIE-Bench dataset \cite{ju2023direct} comprises 700 images, each associated with 10 distinct editing types. Each example includes a source prompt, target prompt, instruction prompt, and source image. In the descriptive setting, only the source and target prompts are used for text guidance, while in the instructive setting, only the instruction prompt is utilized. 

However, the PIE-Bench dataset only provides short text prompts, whereas long, detailed text prompts are required to ensure disentangled edits and prevent artifacts. To ensure a fair comparison in the descriptive setting, we utilize short source and target prompts from the dataset and freeze the attention map \cite{hertz2022prompt} in the first sampling step. In the instructive setting, we employ LLaVA \cite{liu2024visual} to generate a long source caption and adhere to the short instruction from PIE-Bench to obtain a long target prompt as discuss in Section~\ref{sec:instruct}. Our results demonstrate that our method can better adhere to the text guidance and preserve the background compared to current state-of-the-art methods in both descriptive and instructive settings in Table~\ref{tab:comparison} and Supplementary Table~\ref{tab:instruct_comparison}.

\subsection{Qualitative Comparison}
\input{fig/comparison}

\input{fig/instruct}

Our method inherently supports various inversion steps. In the context of single-step inversion, DDIM inversion \cite{song2020denoising} produces a significant number of artifacts, while DDPM inversion \cite{huberman2023edit} generates images with the target attribute but fails to resemble the input image in Supplementary Fig~\ref{fig:ddim}. In contrast, our method successfully generates correct edits that closely resemble the input image, albeit with minor artifacts in the hand and face regions, as well as salt and pepper noise in the image. When considering four-step inversion, all methods exhibit superior results compared to their single-step counterparts. However, both DDIM inversion and DDPM inversion are prone to creating large artifacts when performing substantial structural changes (e.g., transforming a dog into a cat), whereas our method achieves photorealistic edits with significantly higher identity preservation in Supplementary Fig~\ref{fig:ddim}.

Furthermore, we conducted a comparative analysis between our four-step method and an image editing method based on a multi-step diffusion model using descriptive prompt as guidance in Fig~\ref{fig:comparison}. InfEdit \cite{xu2023inversion} and Pix2PixZero \cite{parmar2023zero} distorted the structure of objects such as houses, teddy bears, and guitars. Additionally, Ledits \cite{tsaban2023ledits} and Ledits++ \cite{brack2023ledits++} struggled with large structural changes, such as adding a hat or transforming a man into a woman. In contrast, our method excels in performing realistic edits for both texture and structure changes while maintaining strong identity preservation. Compared to a concurrent work ReNoise \cite{garibi2024renoise} that also relies on SDXL-Turbo \cite{sauer2023adversarial}, we only need 8 NFE per inversion instead of 36 NFE, better preserve the face identity, and produce fewer artifacts in Supplementary Fig~\ref{fig:renoise}.

Similarly, we compare our method with other instruction based methods using instructive prompt as guidance in Fig~\ref{fig:instruct}. Although InstructPix2Pix \cite{brooks2023instructpix2pix} and its variant \cite{hui2024hq,zhang2024magicbrush} need a large scale surprised training set, computation intensive training, and multi-steps sampling, while our inversion network is trained unsupervised by reconstruction loss and only requires four sampling steps, our method outperform them in terms of identity preservation (cat to dog) and text prompt alignment (sweater to T-shirt).

\vspace{-1mm}
\subsection{Ablation Study} 
\vspace{-1mm}
We verify the necessity of each component in our framework through ablation study. First, we visualize the inversion results with varying numbers of inversion steps. Our findings indicate that multi-step inversion is essential for preserving facial identity and preventing blurring artifacts in Supplementary Fig~\ref{fig:inversion_abl}. Subsequently, we calculate the reconstruction metric using 10k validation images, revealing a consistent improvement in reconstruction quality with an increasing number of inversion steps in Supplementary Table~\ref{tab:inversion}. Additionally, we demonstrate that a detailed text prompt condition is crucial for structure preservation and for preventing background artifacts in Supplementary Fig~\ref{fig:long_short_prompt}. Finally, we show local masking is important for preventing background structure change and identity shift in Supplementary Fig~\ref{fig:mask}.

%% file: fig/comparison.tex
\begin{figure}[tb]
	\centering
	\setlength{\tabcolsep}{1pt}	
	\begin{tabular}{ccccccc}

		& {\scriptsize Original} &{\scriptsize Ours} & {\scriptsize InfEdit} & {\scriptsize Pix2PixZero} & {\scriptsize Ledits} &{\scriptsize Ledits++} \\
		\rotatebox{90}{ \scriptsize  Dog $\rightarrow$ Husky} &
            
		\includegraphics[width=0.16\columnwidth]{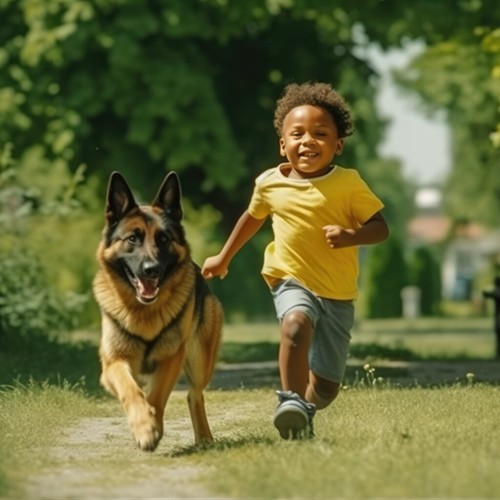} &
            \includegraphics[width=0.16\columnwidth]{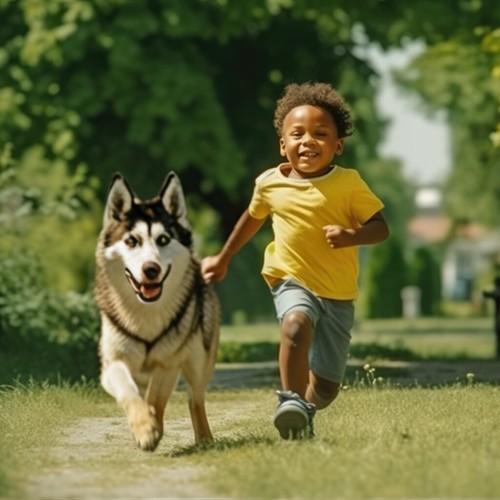} &
            \includegraphics[width=0.16\columnwidth]{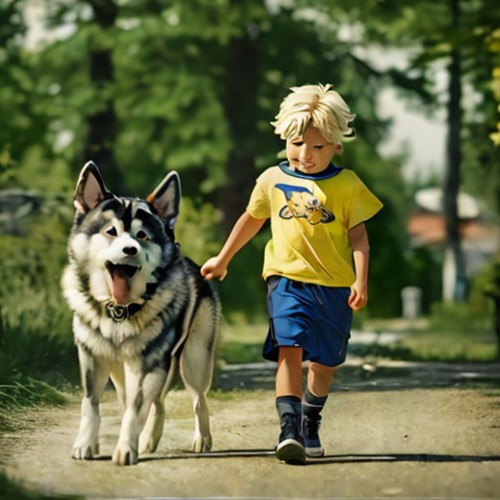} &
            \includegraphics[width=0.16\columnwidth]{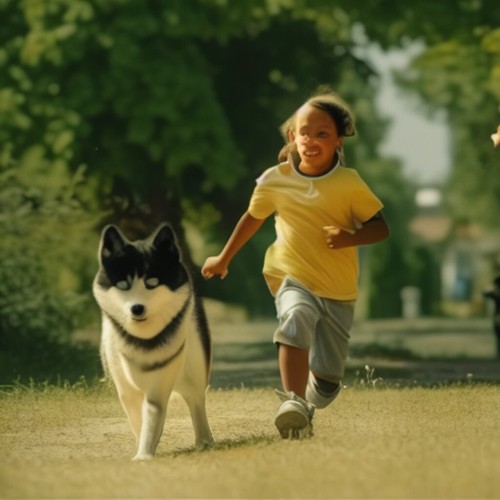} &
            \includegraphics[width=0.16\columnwidth]{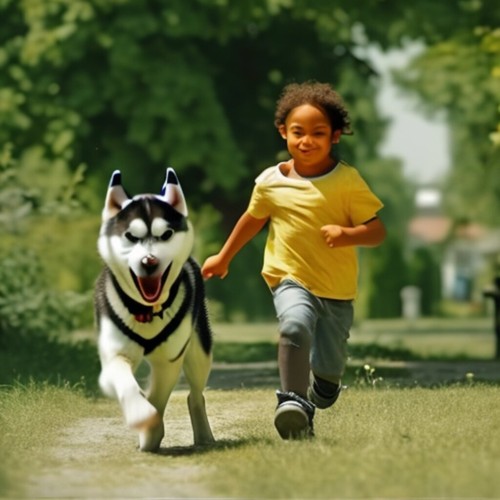} &
            \includegraphics[width=0.16\columnwidth]{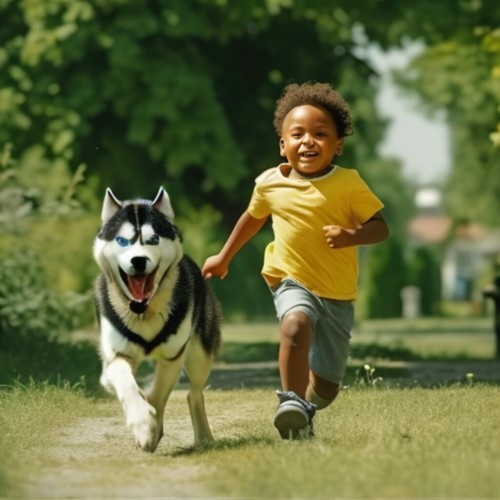} 
		\\
  
  		\rotatebox{90}{ \scriptsize \phantom{kk} In Fall} &
		\includegraphics[width=0.16\columnwidth]{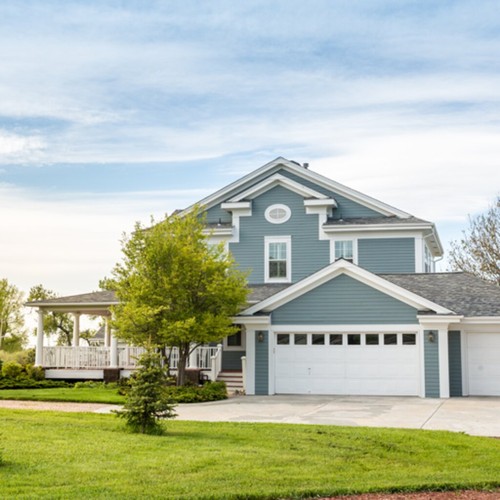} &
            \includegraphics[width=0.16\columnwidth]{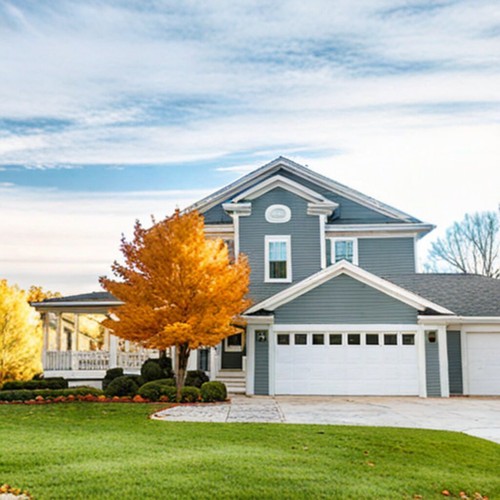} &
            \includegraphics[width=0.16\columnwidth]{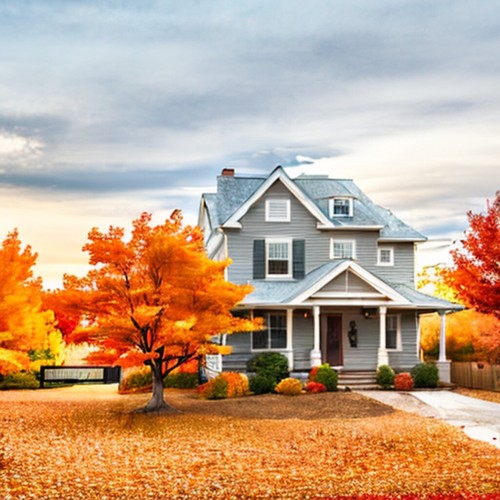} &
            \includegraphics[width=0.16\columnwidth]{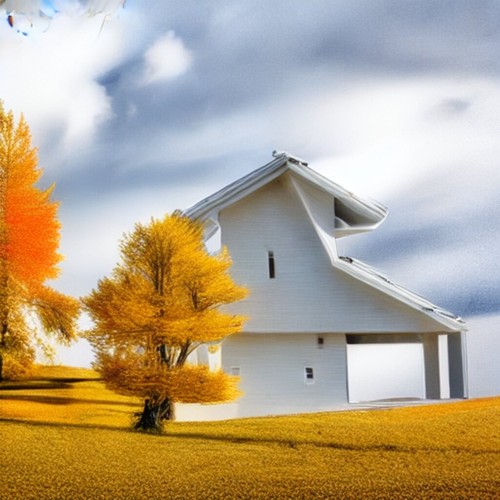} &
            \includegraphics[width=0.16\columnwidth]{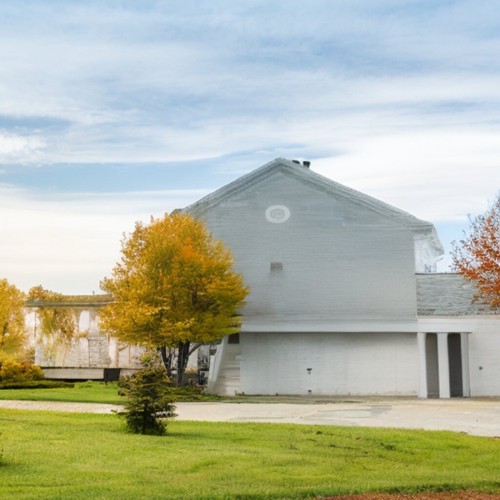} &
            \includegraphics[width=0.16\columnwidth]{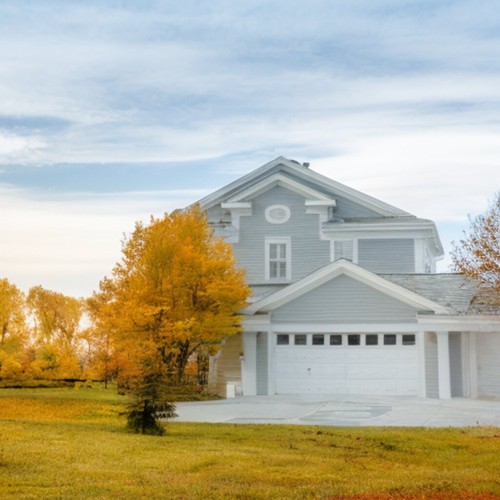} 
		\\
  
  		\rotatebox{90}{ \scriptsize \phantom{k} + Top Hat} &
		\includegraphics[width=0.16\columnwidth]{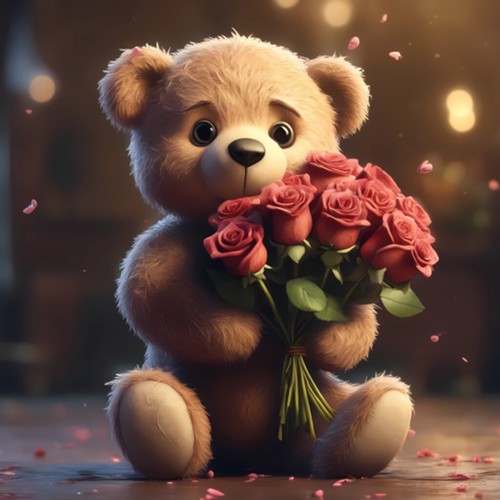} &
            \includegraphics[width=0.16\columnwidth]{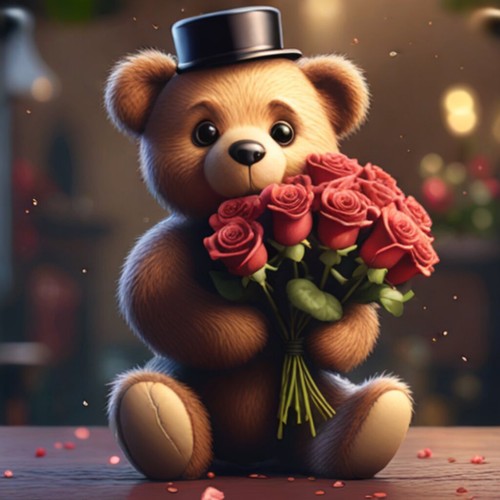} &
            \includegraphics[width=0.16\columnwidth]{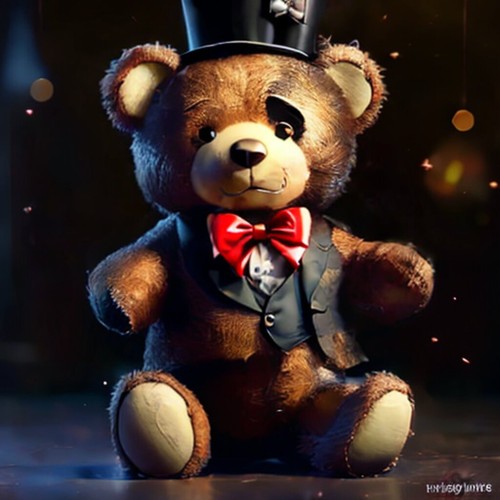} &
            \includegraphics[width=0.16\columnwidth]{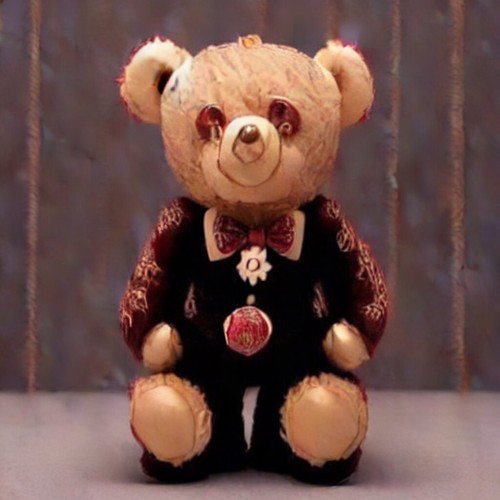} &
            \includegraphics[width=0.16\columnwidth]{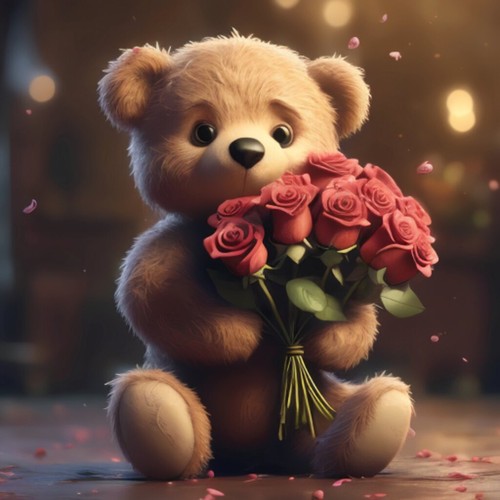} &
            \includegraphics[width=0.16\columnwidth]{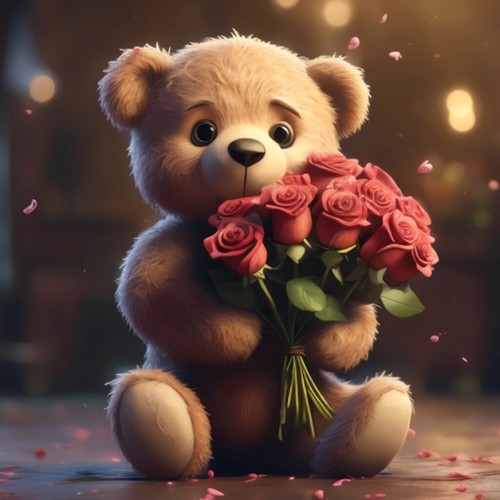} 
		\\
  
  		\rotatebox{90}{ \scriptsize Woman$\rightarrow$Man} &
		\includegraphics[width=0.16\columnwidth]{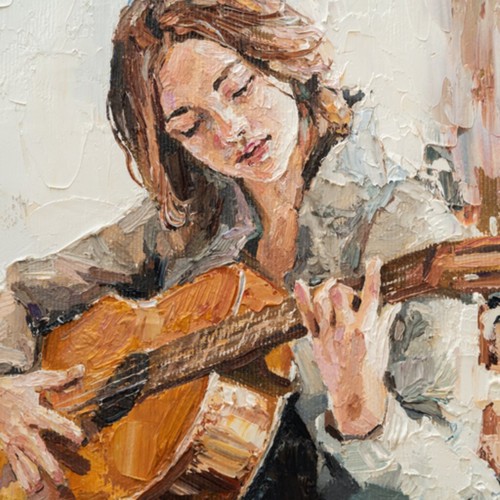} &
            \includegraphics[width=0.16\columnwidth]{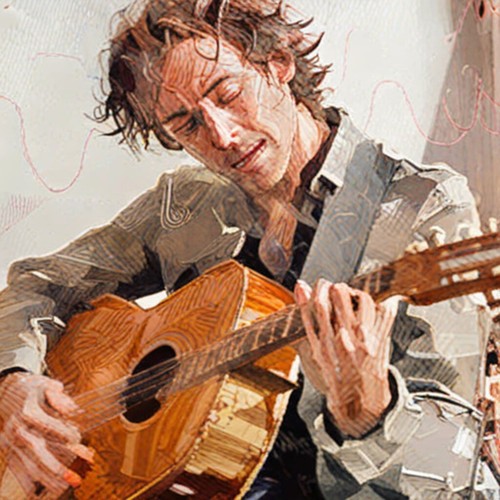} &
            \includegraphics[width=0.16\columnwidth]{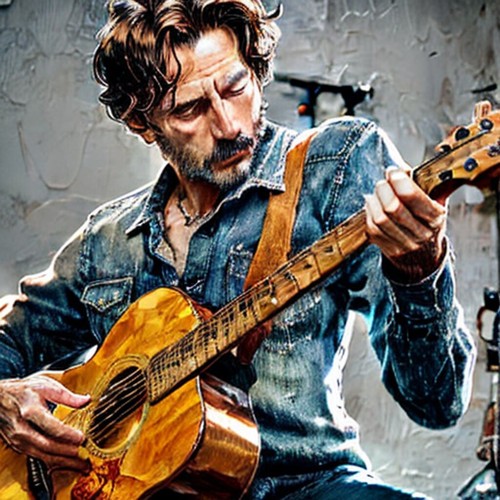} &
            \includegraphics[width=0.16\columnwidth]{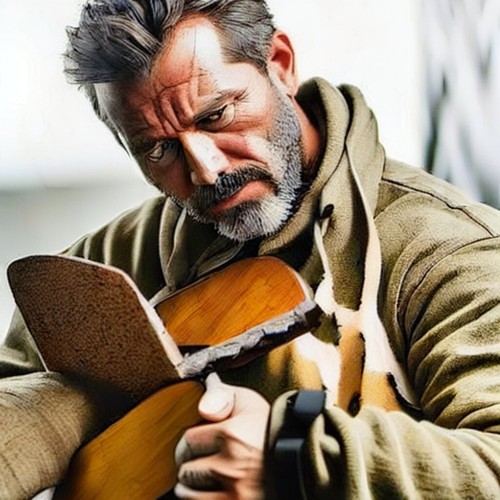} &
            \includegraphics[width=0.16\columnwidth]{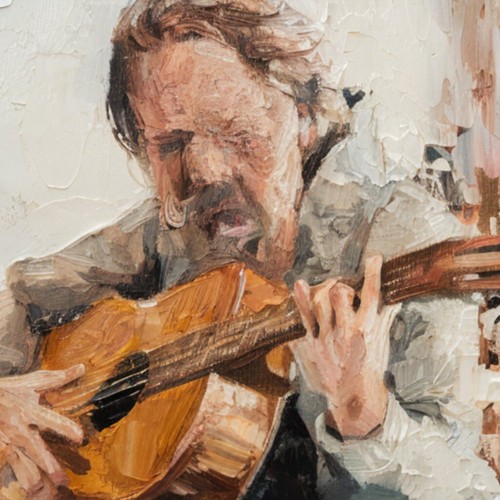} &
            \includegraphics[width=0.16\columnwidth]{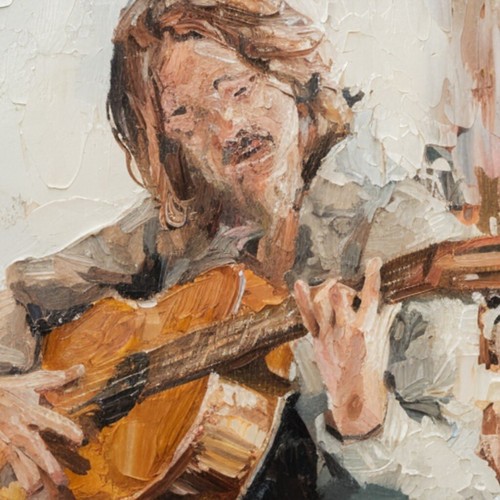} 
		\\

	\end{tabular}
 \vspace{-3mm}
	\caption{We compare methods using descriptive text prompt as guidance. Despite requiring only four steps, our method outperforms multi-step methods, particularly in scenarios requiring significant structural changes for attributes such as adding a hat or transforming a man into a woman. In contrast, InfEdit and Pix2PixZero struggle with background and identity preservation. Similarly, Ledits and Ledits++ are unable to effectively handle large structural changes, as evidenced by their failure in adding a top hat or transforming a man into a woman.}
	\label{fig:comparison}
  \vspace{-5mm}
\end{figure}

%% file: fig/instruct.tex
\begin{figure}[tb]
	\centering
	\setlength{\tabcolsep}{1pt}	
	\begin{tabular}{ccccccc}

		 {\scriptsize Original} &{\scriptsize Ours} & {\scriptsize MagicBrush} & {\scriptsize HQEdit} & {\scriptsize InstructPix2Pix} \\

		\includegraphics[width=0.18\columnwidth]{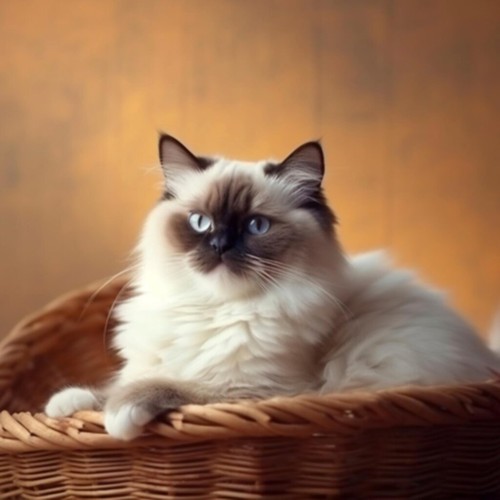} &
            \includegraphics[width=0.18\columnwidth]{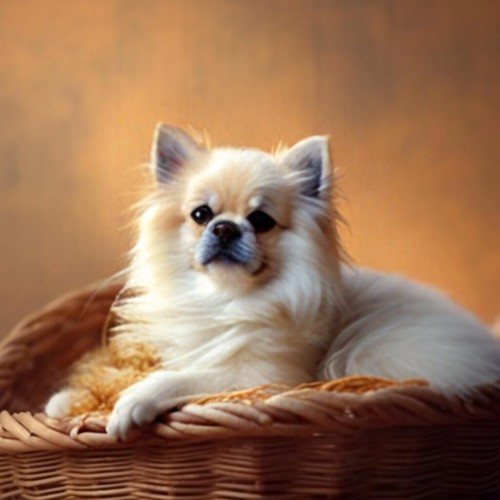} &
            \includegraphics[width=0.18\columnwidth]{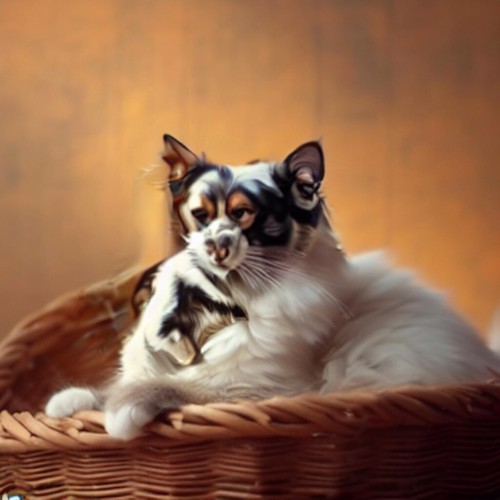} &
            \includegraphics[width=0.18\columnwidth]{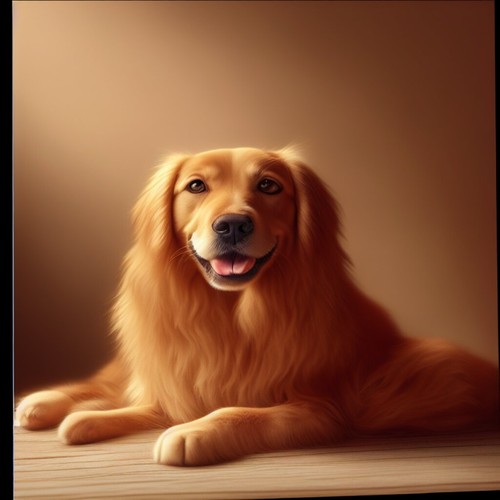} &
            \includegraphics[width=0.18\columnwidth]{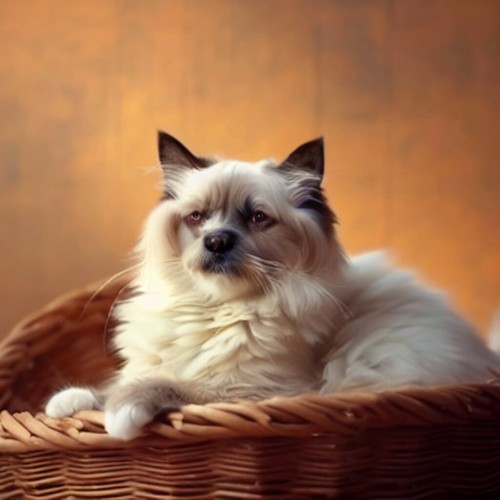} 
		\\
            \multicolumn{5}{c}{\scriptsize Replace the cat with a dog}\\ 

		\includegraphics[width=0.18\columnwidth]{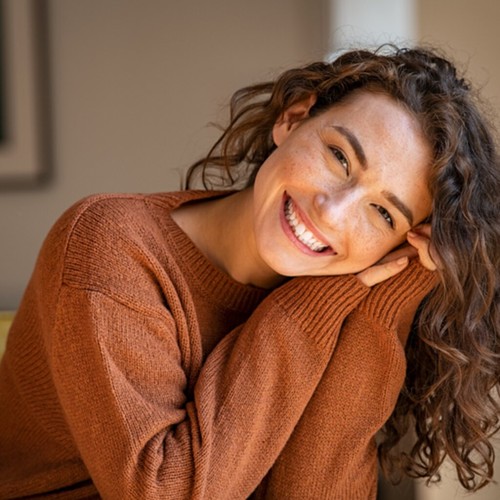} &
            \includegraphics[width=0.18\columnwidth]{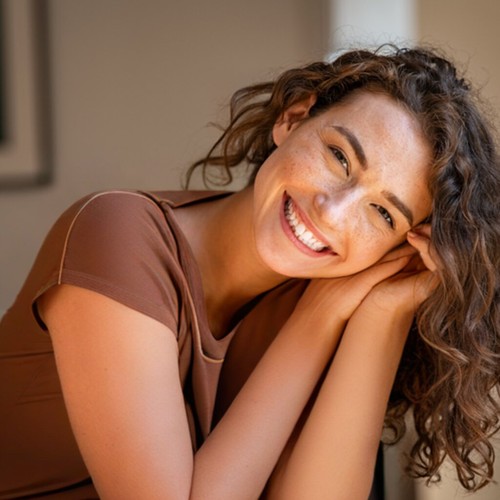} &
            \includegraphics[width=0.18\columnwidth]{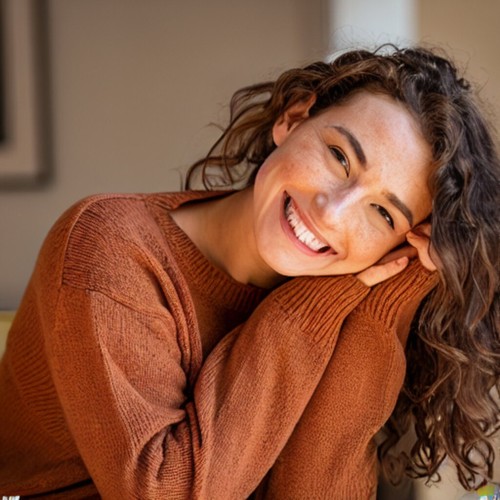} &
            \includegraphics[width=0.18\columnwidth]{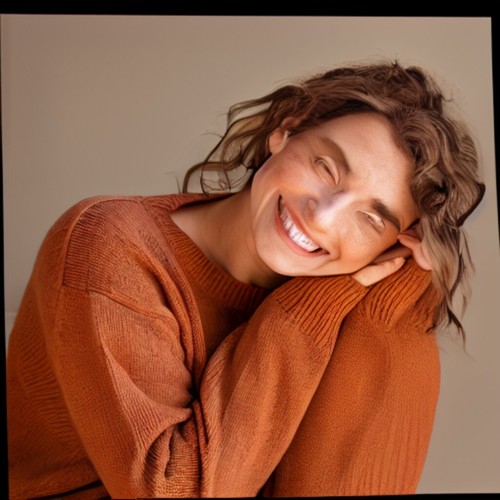} &
            \includegraphics[width=0.18\columnwidth]{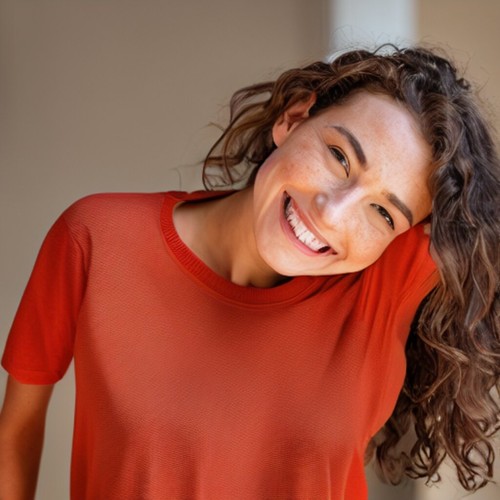} 
		\\
            \multicolumn{5}{c}{\scriptsize Change the sweater to T-shirt}\\

		\includegraphics[width=0.18\columnwidth]{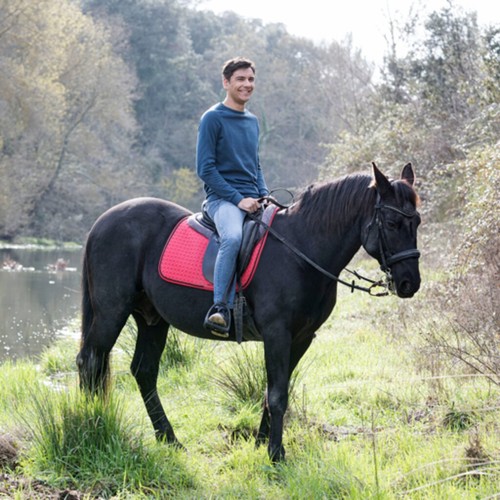} &
            \includegraphics[width=0.18\columnwidth]{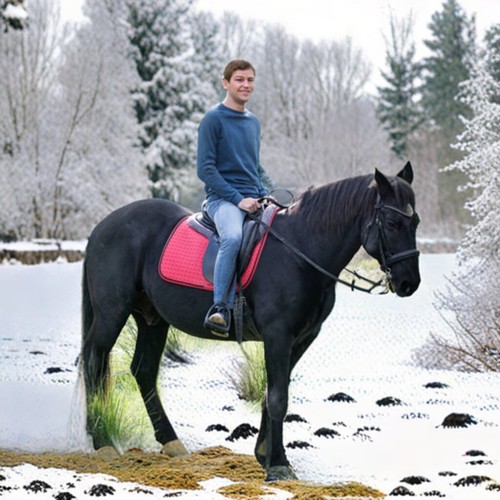} &
            \includegraphics[width=0.18\columnwidth]{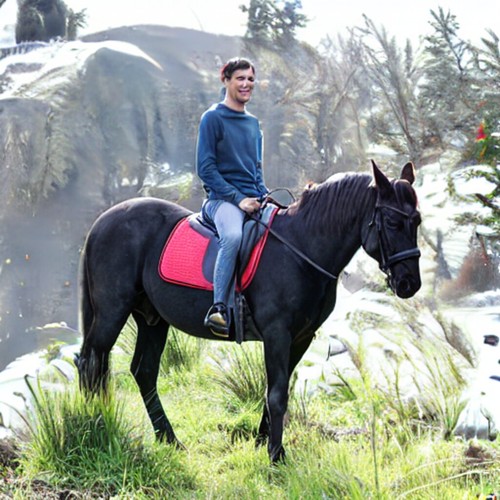} &
            \includegraphics[width=0.18\columnwidth]{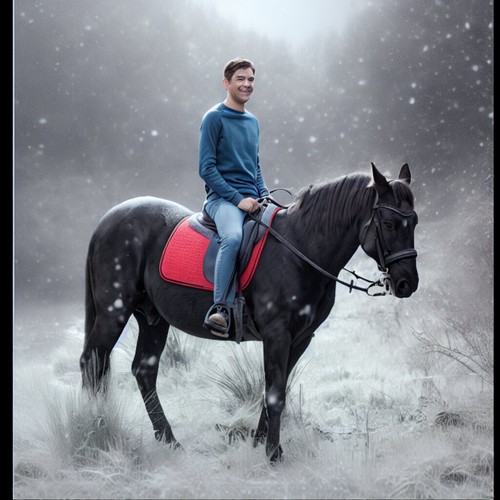} &
            \includegraphics[width=0.18\columnwidth]{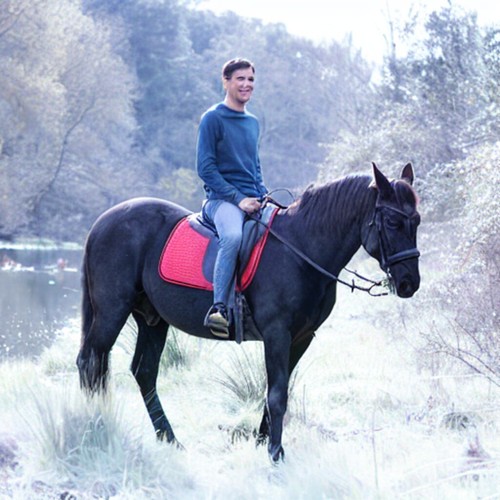} 
		\\
            \multicolumn{5}{c}{\scriptsize Add snow to the scene}\\

	\end{tabular}
        \vspace{-2mm}
	\caption{We evaluate methods utilizing instructive prompts as guidance. Although our method does not require any surprised training and only requires four sampling steps, it outperforms InstructPix2Pix and its variants, in terms of identity preservation (cat to dog) and text prompt alignment (sweater to T-shirt). It is worth to mention InstructPix2pIx and its variants require collecting a large scale surprised training set, computation intensive training, and multi-steps sampling.}
	\label{fig:instruct}
  \vspace{-8mm}
\end{figure}

%% file: 5_limitations.tex
\vspace{-2mm}
\section{Limitations and Societal Impact}
\vspace{-2mm}

First, our method relies on LLaVA \cite{liu2024visual} to generate detailed captions. However, as we only perform a few-step inversion, the computationally intensive LLaVA model becomes a bottleneck. Hence, there is a need to explore alternative lightweight caption models to enable real-time image inversion. Secondly, while attention masks can effectively confine the editing region, they are often imprecise and may encompass nearby regions, which cannot be fully addressed by increasing the attention threshold. This imprecision can lead to slight identity shifts, particularly when the editing region is near a human face. We demonstrate that this issue can be mitigated by using a rough user-provided mask in Supplementary Fig~\ref{fig:manual_mask}. Lastly, our method is not able perform large pose change (turning a running man to a sitting man) in Supplementary Fig~\ref{fig:failure}.

Our method, as a generative image editing tool, offers both creative opportunities and challenges. While it enables innovative image editing capability, it also raises concerns about the creation and dissemination of manipulated data, misinformation, and spam. One notable issue is the rise of deliberate image manipulation, commonly known as "deep fakes", which disproportionately affects women.

%% file: 6_conclusion.tex
\vspace{-2mm}
\section{Conclusion}
\vspace{-1mm}

To the best of our knowledge, our method is the first work exploring image editing in the context of few-step diffusion models, and also the first to explore encoder based inversion in diffusion model.  
We demonstrate that disentangled controls can be easily achieved in the few-step diffusion model by conditioning on an (automatically generated) detailed text prompt. Our method enables users to make realistic text-guided image edits at interactive rates, running in milliseconds for both the inversion and editing processes.

%% file: 7_sup.tex
\input{fig/multi_attribute}

\input{fig/ddim}

\input{fig/attention}

\input{fig/long_short_prompt}

\input{fig/inversion_abl}

\input{fig/mask}

\input{fig/manual_mask}

\input{fig/renoise}
\input{fig/failure}

\begin{table}
\scriptsize
\centering
\resizebox{.8\textwidth}{!}{
  \begin{tabular}{cccccccc}
  \toprule
method & steps &{L2{$^\downarrow_{10^{-3}}$}} & PSNR $\uparrow$ & LPIPS $\downarrow$ & FID $\downarrow$ & {KID{$^\downarrow_{10^{-4}}$}}  \\
\hline
    &1 & 47.9 & 13.8 & 0.534 & 56.9  & 111.0\\
ddim &2  & 39.9 & 14.8  & 0.444 & 51.8 & 77.9 \\
    &4  & 29.1  & 16.2 & 0.363 & 41.7 & 42.2\\
 \hline
    &1 & 20.8 & 17.5 & 0.349 &37.7  & 45.1\\
ours &2  & 8.90 & 21.7  & 0.220 & 24.9 & 10.5 \\
    &4  & \textbf{5.87}  & \textbf{23.7} & \textbf{0.163} & \textbf{18.7} & \textbf{4.33}\\
    \bottomrule
  \end{tabular}
  }
  \caption{ 
  Image reconstruction quality consistently improves with an increasing number of inversion steps. The evaluation metrics are computed over a validation set comprising 10,000 images. The LPIPS (Learned Perceptual Image Patch Similarity) loss is calculated using an AlexNet \cite{krizhevsky2017imagenet} backbone, following established practices in the field \cite{zhang2018unreasonable}. }
  \label{tab:inversion}
\end{table}

\begin{table}
  \centering
  \begin{tabular}{l|llll|ll}

    \toprule
    Method &
      \multicolumn{4}{c}{Background Preservation} &
      \multicolumn{2}{c}{CLIP Similarity}  \\
        \cline{2-7} %
      &  {PSNR $\uparrow$ } & {LPIPS{$^\downarrow_{10^3}$}} & {MSE{$^\downarrow_{10^4}$}} & {SSIM{$^\uparrow_{10^2}$}} &  {Whole $\uparrow$ } & {Edited $\uparrow$} \\
      
    InstructPix2Pix & 20.58 & 171.51 & 305.47 & 74.82 & 23.43 & \textbf{21.58}\\
    HQEdit & 12.47 & 308.77 & 793.64 & 57.43 & 22.46 &21.24\\
    MagicBrush & 27.36 & 64.21 & 121.54 & 84.02 & 23.24 & 20.58\\
    Ours (\textit{instructive}) &  \textbf{32.63} & \textbf{33.88} & \textbf{17.95} & \textbf{93.57} & \textbf{23.78} & 20.97\\
    \bottomrule
  \end{tabular}
  \caption{ Image editing comparison using instructive text in PIE-Bench dataset. We only use instructive prompt in this comparison. Even though our model is not trained on the instructive setting, it outperforms the instructive training models by a large margins in background preservation, and achieves similar scores in CLIP similarity.}
  \label{tab:instruct_comparison}
\end{table}

%% file: fig/multi_attribute.tex
\begin{figure}[!htb]
	\centering
	\setlength{\tabcolsep}{1pt}	
	\begin{tabular}{cccccc}

		   & \multirow{2}{4em}{\tiny {Shirt$\rightarrow$Suit} +Glasses } & & \multirow{2}{4em}{\tiny {Dog $\rightarrow$ Cat} {boy $\rightarrow$ Girl}} & & \multirow{2}{4em}{\tiny Car $\rightarrow$ SUV {+Winter}}  \\
     \\
		\includegraphics[width=0.16\columnwidth]{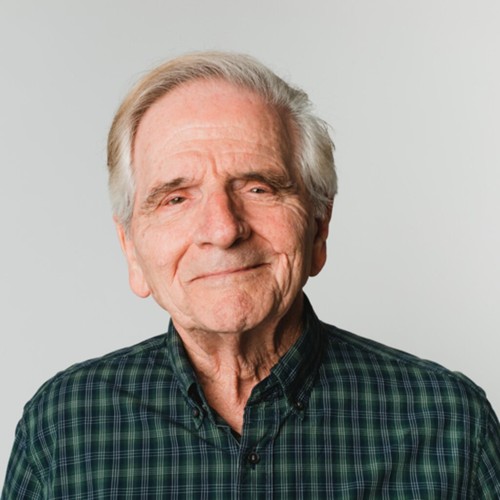} &
            \includegraphics[width=0.16\columnwidth]{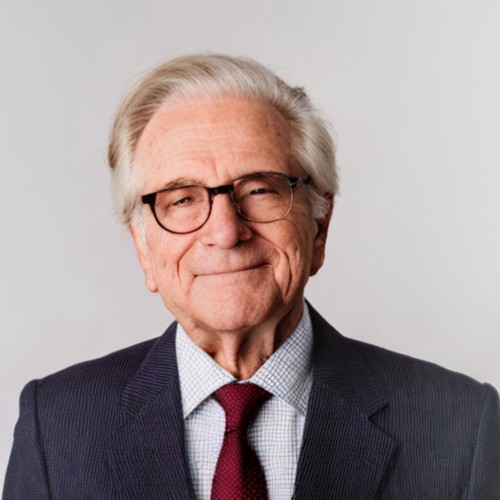} &
		\includegraphics[width=0.16\columnwidth]{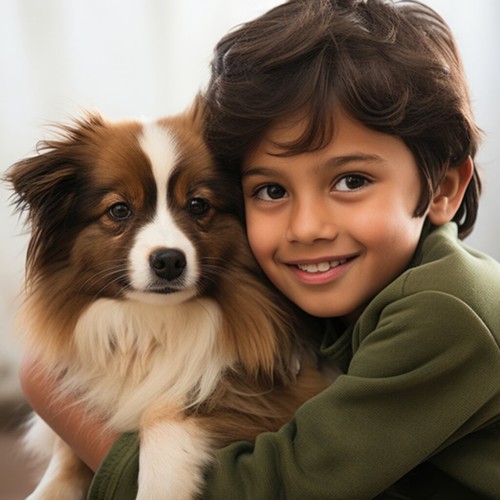} &
            \includegraphics[width=0.16\columnwidth]{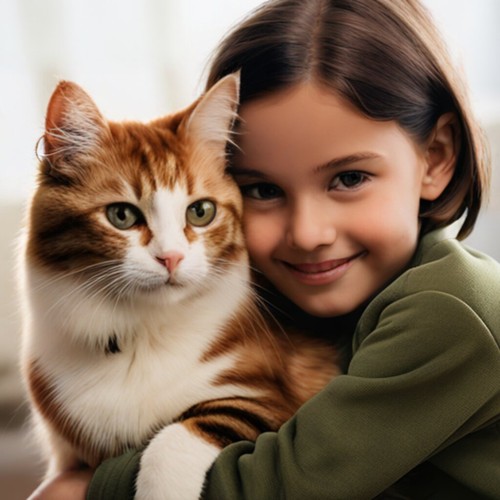} &
		\includegraphics[width=0.16\columnwidth]{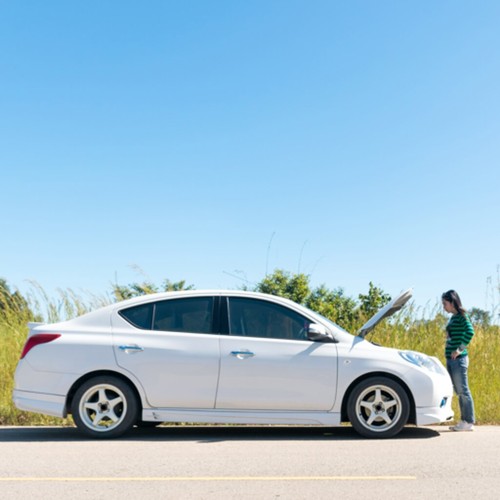} &
            \includegraphics[width=0.16\columnwidth]{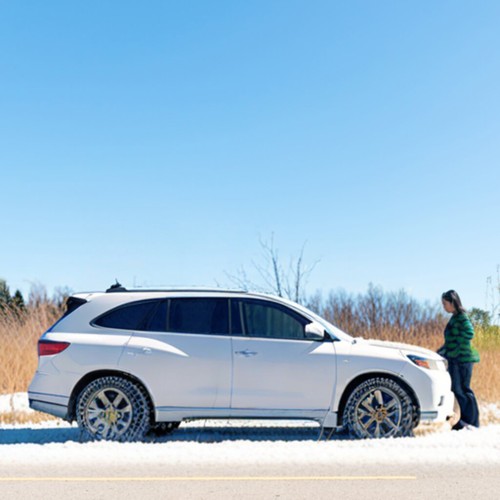} 
		\\
 		& \tiny Straight Hair  & \tiny Black Hair & \tiny Sweater $\rightarrow$ Suit & \tiny + Sunglasses & \tiny Image $\rightarrow$ Drawing \\ 
  		\includegraphics[width=0.16\columnwidth]{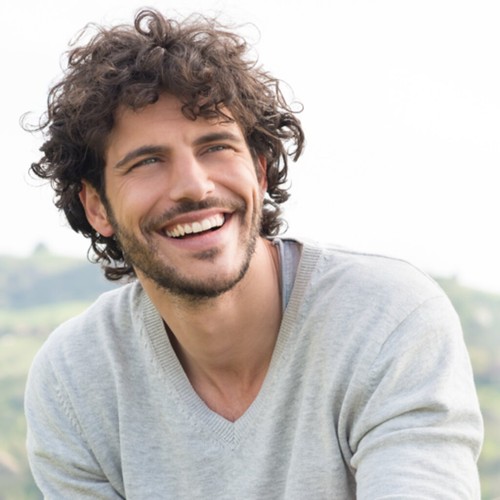} &
            \includegraphics[width=0.16\columnwidth]{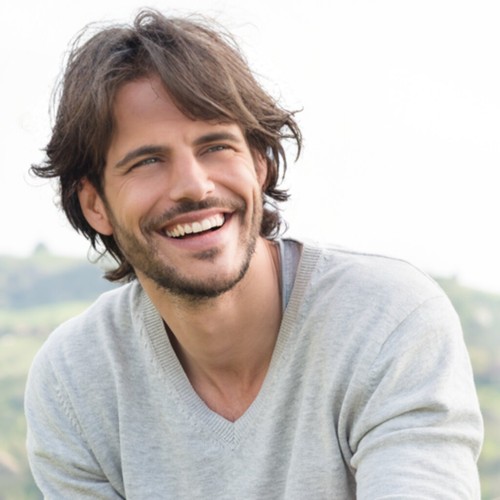} &
		\includegraphics[width=0.16\columnwidth]{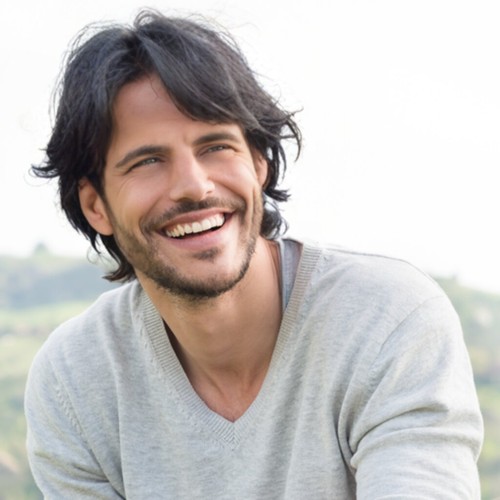} &
            \includegraphics[width=0.16\columnwidth]{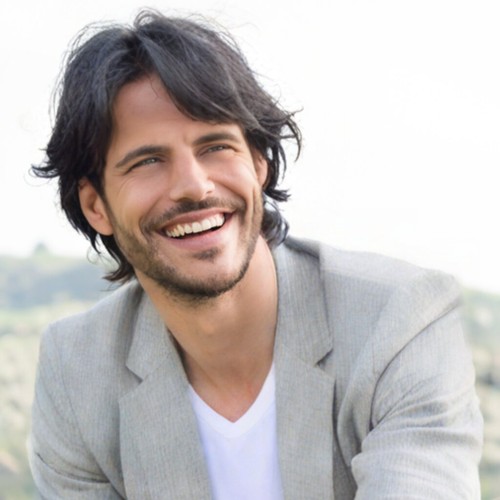} &
		\includegraphics[width=0.16\columnwidth]{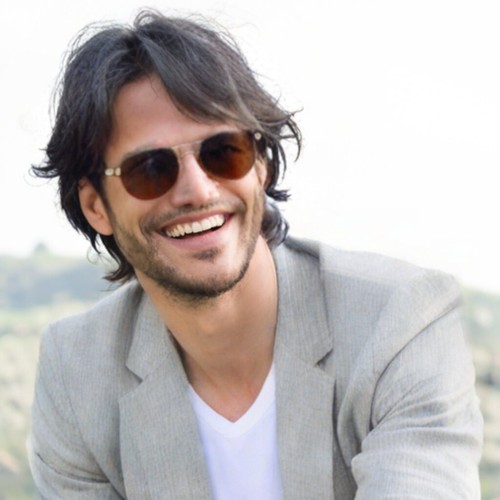} &
            \includegraphics[width=0.16\columnwidth]{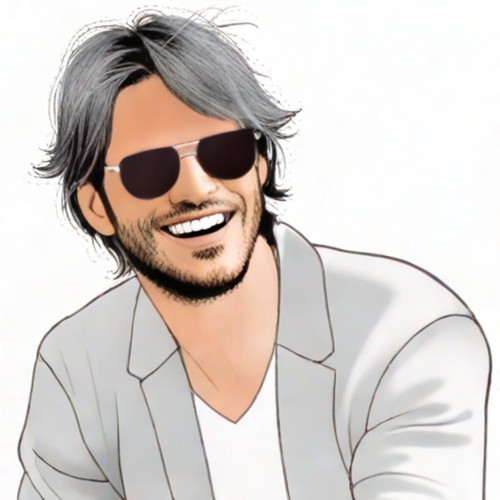} 
		\\
	\end{tabular}
 
	\caption{ Our method can manipulate multiple attributes simultaneously (in the first row), or perform continuous editing (in the second row).  
    }
	\label{fig:multi_attribute}
\end{figure}

%% file: fig/ddim.tex
\begin{figure}[tb]
	\centering
	\setlength{\tabcolsep}{1pt}	
	\begin{tabular}{ccccc|ccc}
		   &  &{\footnotesize Ours } & {\footnotesize DDIM}  & {\footnotesize DDPM} &{\footnotesize Ours } & {\footnotesize DDIM}  & {\footnotesize DDPM} \\

            \rotatebox{90}{\scriptsize \phantom{k} Dog$\rightarrow$Cat} &
            \includegraphics[width=0.13\columnwidth]{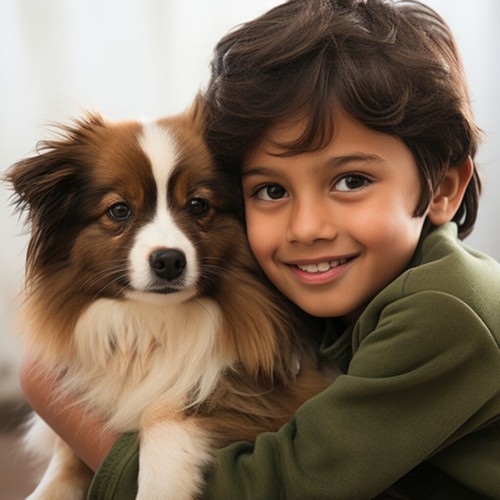} &
            \includegraphics[width=0.13\columnwidth]{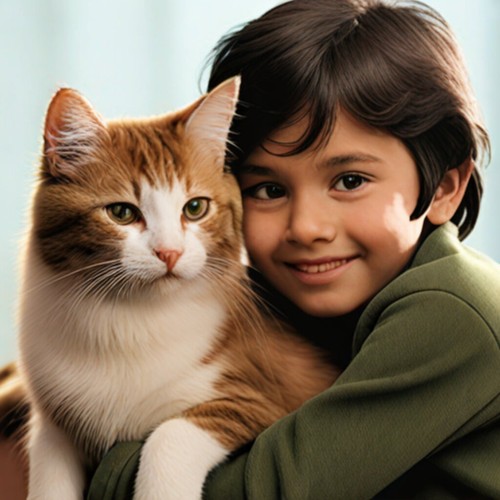} &
            \includegraphics[width=0.13\columnwidth]{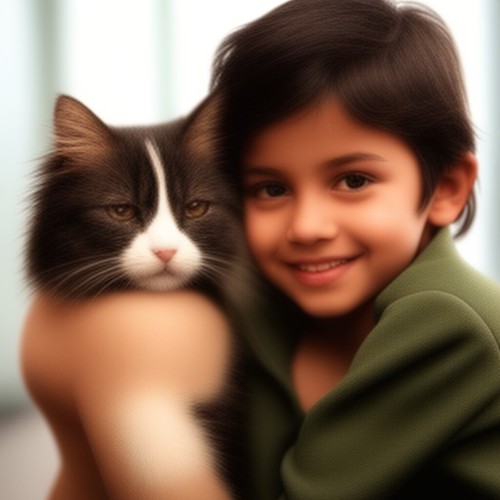} &
            \includegraphics[width=0.13\columnwidth]{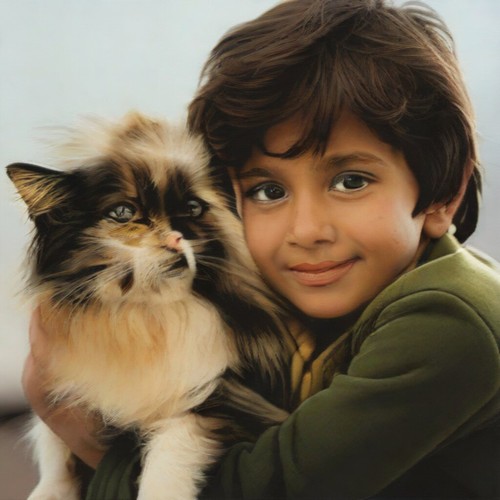} &
            \includegraphics[width=0.13\columnwidth]{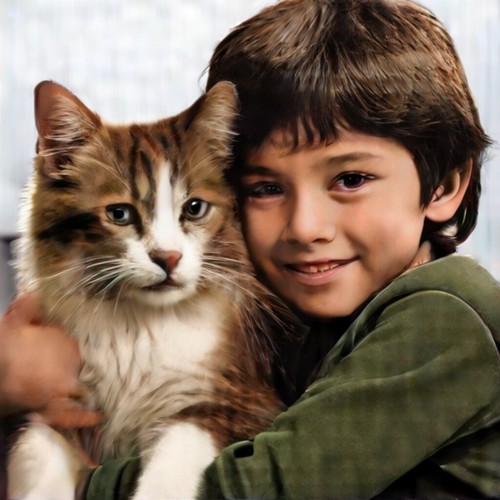} &
            \includegraphics[width=0.13\columnwidth]{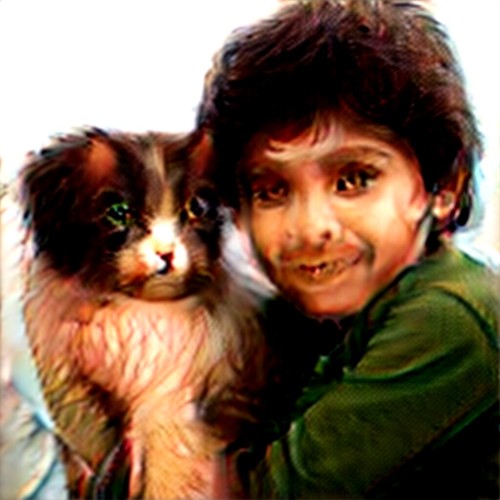} &
            \includegraphics[width=0.13\columnwidth]{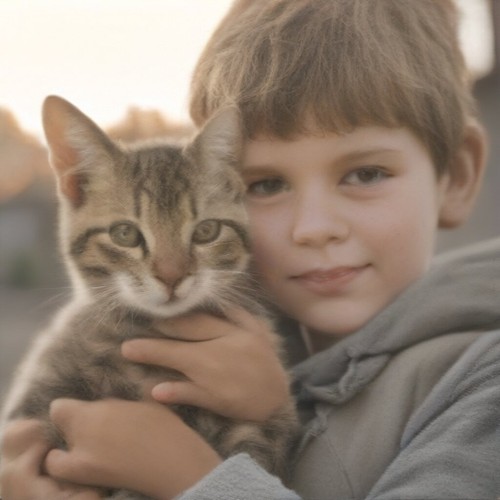} 
		\\
            \rotatebox{90}{\scriptsize \phantom{} Young$\rightarrow$Old} &
            \includegraphics[width=0.13\columnwidth]{./img/ddim/original.jpg} &
            \includegraphics[width=0.13\columnwidth]{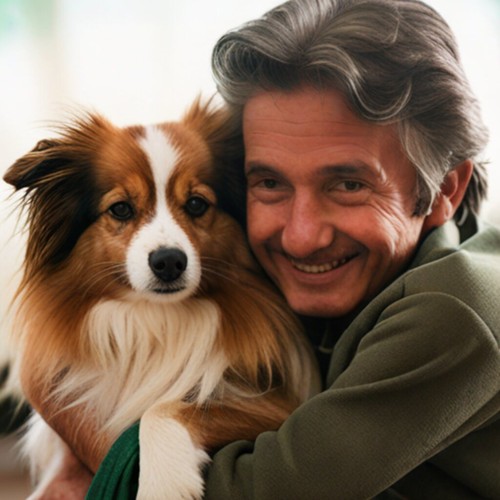} &
            \includegraphics[width=0.13\columnwidth]{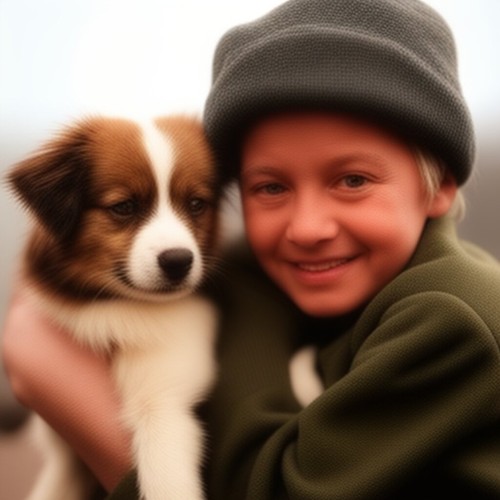} &
            \includegraphics[width=0.13\columnwidth]{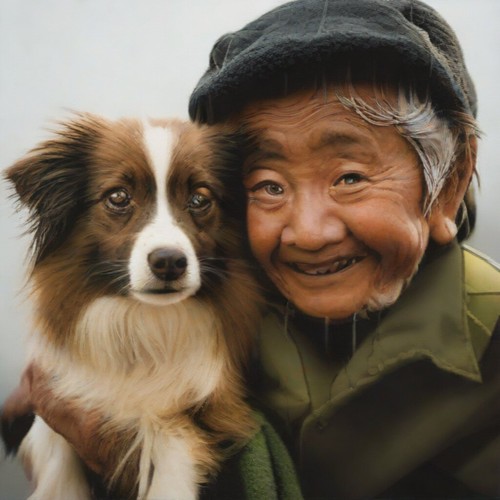} &
            \includegraphics[width=0.13\columnwidth]{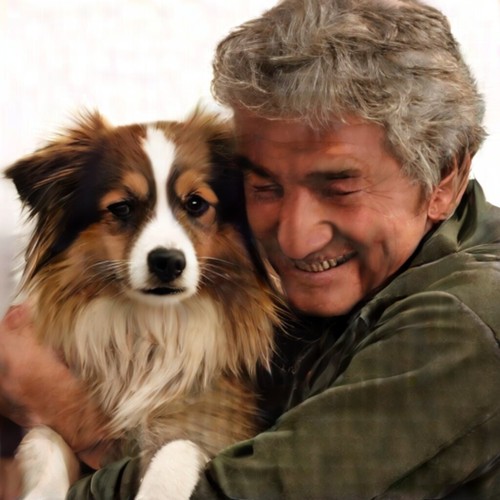} &
            \includegraphics[width=0.13\columnwidth]{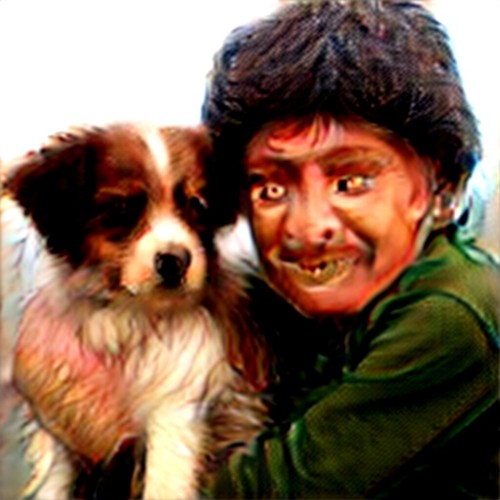} &
            \includegraphics[width=0.13\columnwidth]{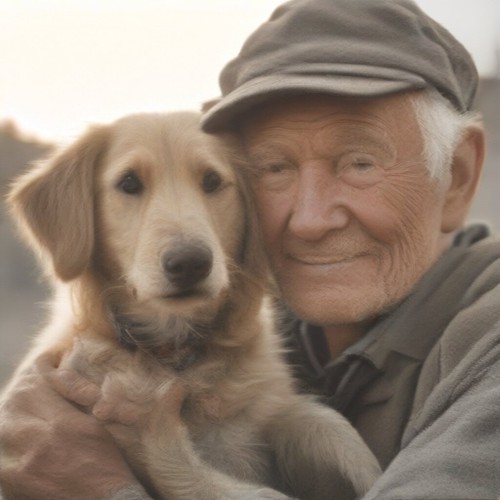} 
		\\
            \rotatebox{90}{\tiny \phantom{k} Shirt$\rightarrow$Jacket} &
            \includegraphics[width=0.13\columnwidth]{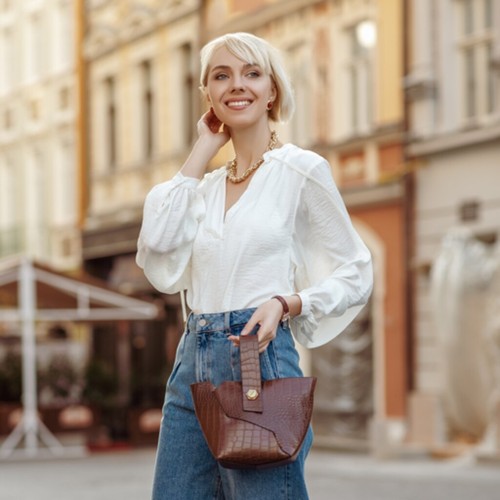} &
            \includegraphics[width=0.13\columnwidth]{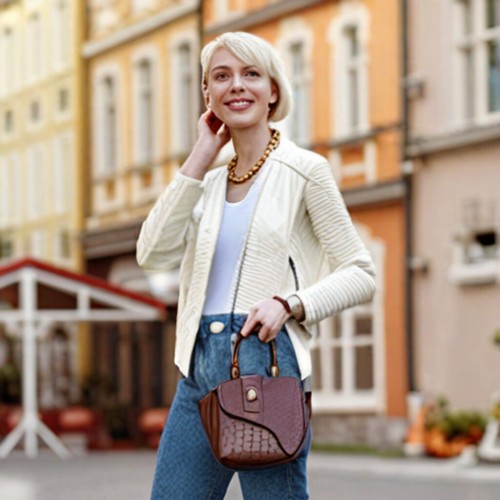} &
            \includegraphics[width=0.13\columnwidth]{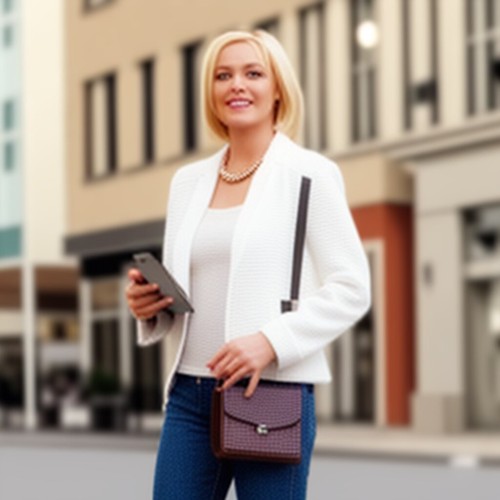} &
            \includegraphics[width=0.13\columnwidth]{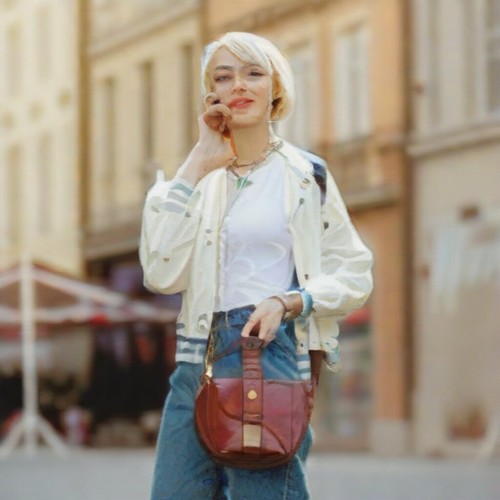} &
            \includegraphics[width=0.13\columnwidth]{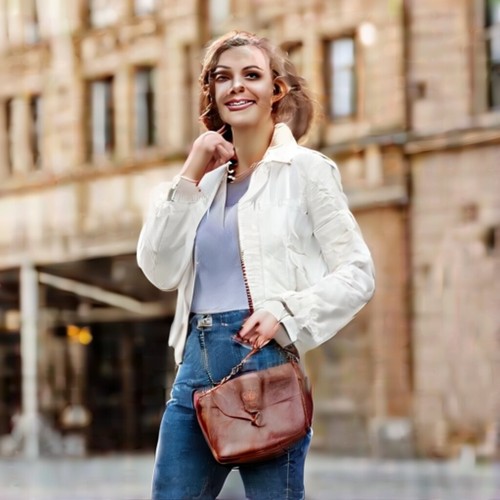} &
            \includegraphics[width=0.13\columnwidth]{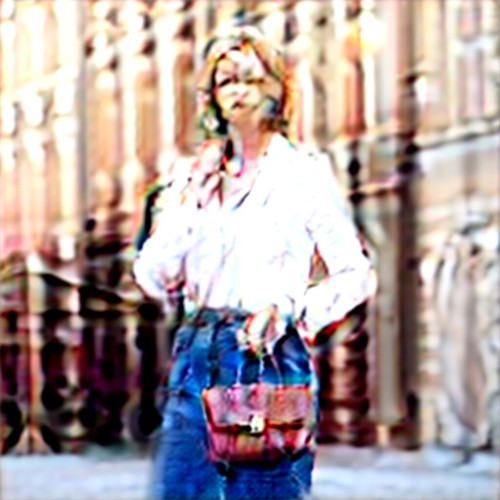} &
            \includegraphics[width=0.13\columnwidth]{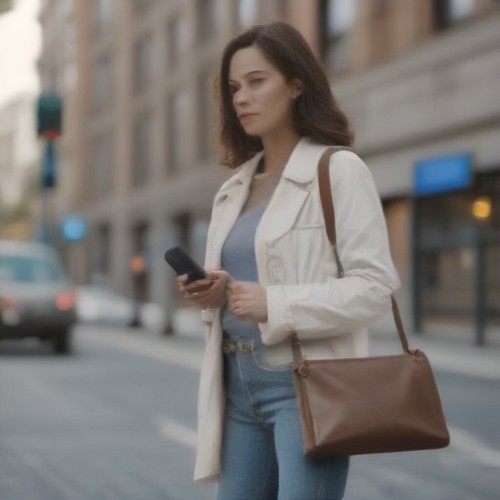} 
		\\
            \rotatebox{90}{\tiny \phantom{} Building$\rightarrow$Lake} &
            \includegraphics[width=0.13\columnwidth]{./img/ddim/woman.jpg} &
            \includegraphics[width=0.13\columnwidth]{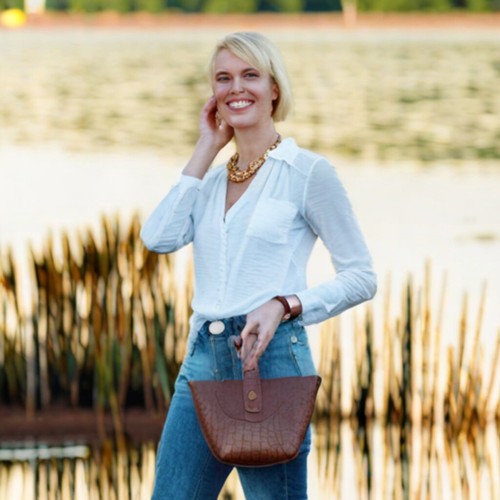} &
            \includegraphics[width=0.13\columnwidth]{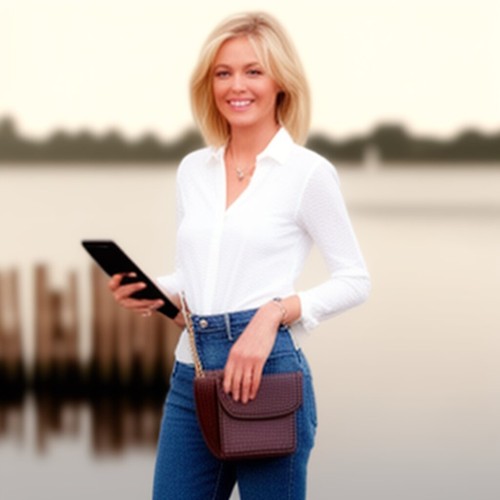} &
            \includegraphics[width=0.13\columnwidth]{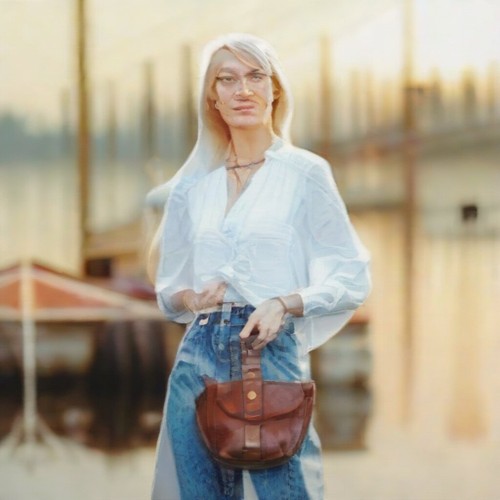} &
            \includegraphics[width=0.13\columnwidth]{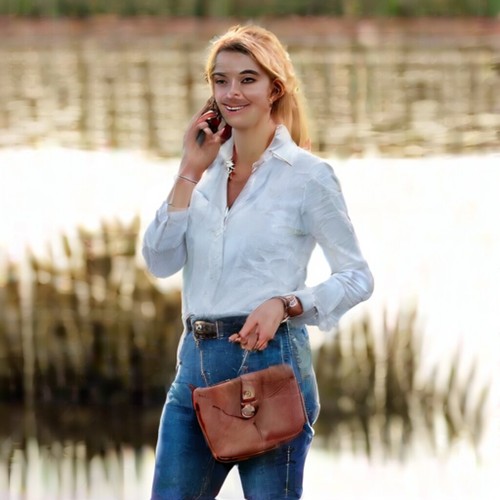} &
            \includegraphics[width=0.13\columnwidth]{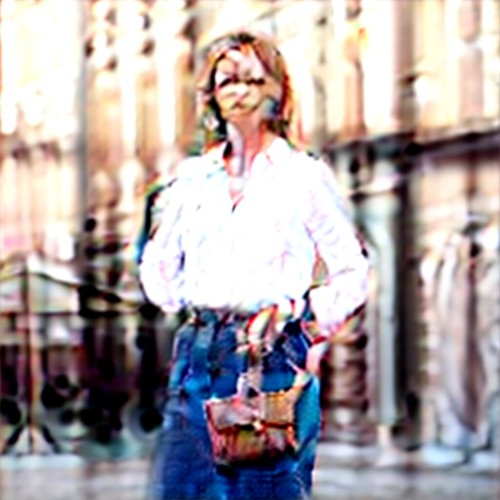} &
            \includegraphics[width=0.13\columnwidth]{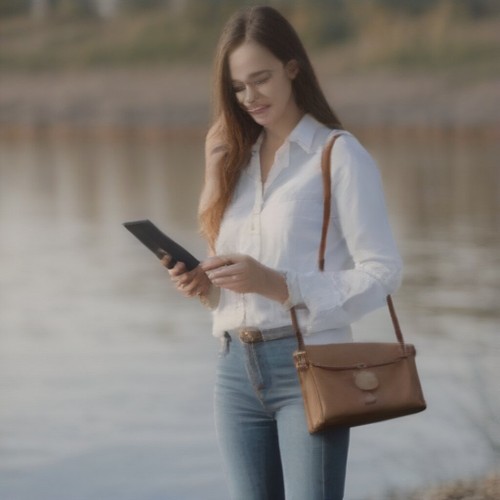} 
		\\
	&& \multicolumn{3}{c}{\ruleline{0.30\textwidth}{4 steps}} &\multicolumn{3}{c}{\ruleline{0.30\textwidth}{1 step}}

	\end{tabular}
	\caption{ Utilizing SDXL Turbo as the foundational model, we conduct a comparative analysis of DDIM inversion, DDPM inversion, and our method. While both DDIM and DDPM inversions completely fail in single step inversion, struggle to effect substantial structural changes (young2old) and tend to introduce pronounced artifacts (dog2cat) in 4 steps inversion. In contrast, our method performs disentangled edit even in a single step, and produce photo realistic edited images in 4 steps.  }
	\label{fig:ddim}
\end{figure}

%% file: fig/attention.tex
\begin{figure}[tb]
	\centering
	\setlength{\tabcolsep}{1pt}	
	\begin{tabular}{cccccccc}
		 & {\footnotesize } & {\footnotesize P2P} & {\footnotesize Ours} & &{\footnotesize } & {\footnotesize P2P} & {\footnotesize Ours}  \\
            \rotatebox{90}{\tiny Horse $\rightarrow$ Unicorn} &
		\includegraphics[width=0.15\columnwidth]{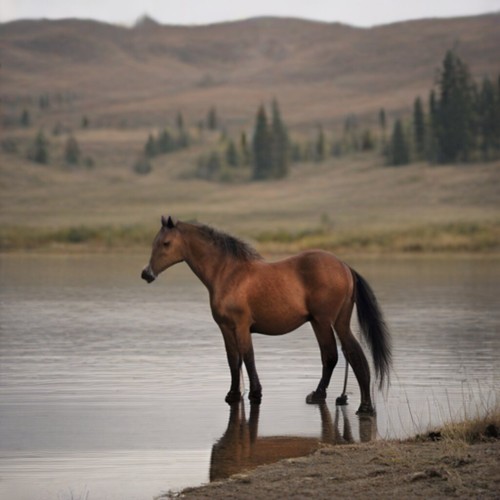} &
		\includegraphics[width=0.15\columnwidth]{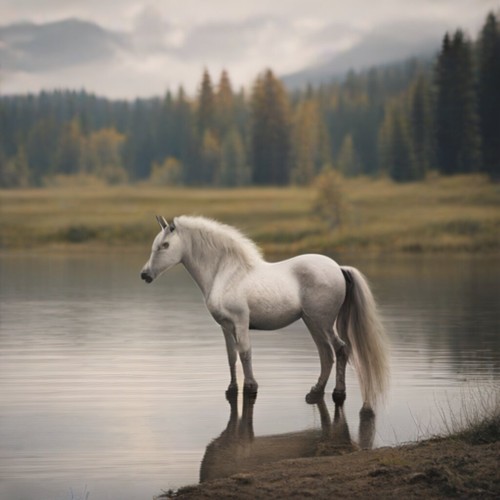} &
            \includegraphics[width=0.15\columnwidth]{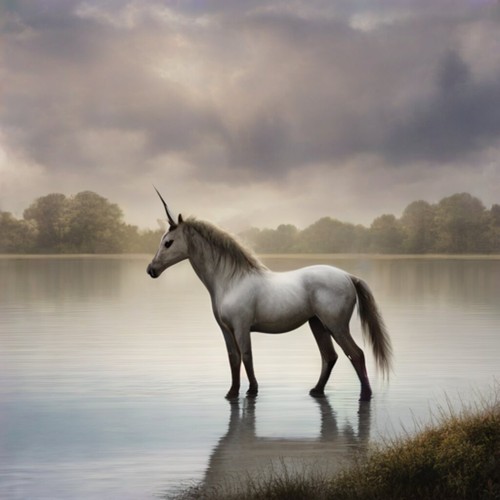}&
		\rotatebox{90}{\tiny \phantom{kk} Fox $\rightarrow$ Dog} &
		\includegraphics[width=0.15\columnwidth]{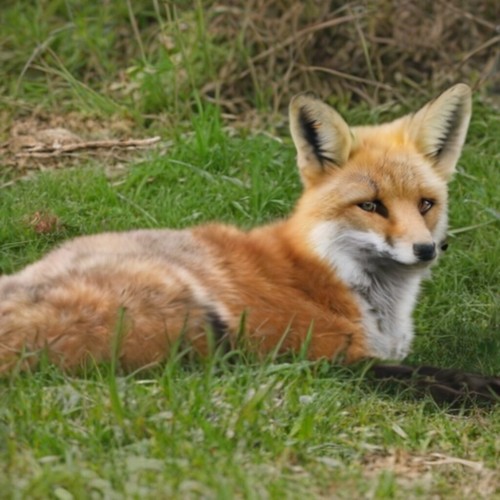} &
		\includegraphics[width=0.15\columnwidth]{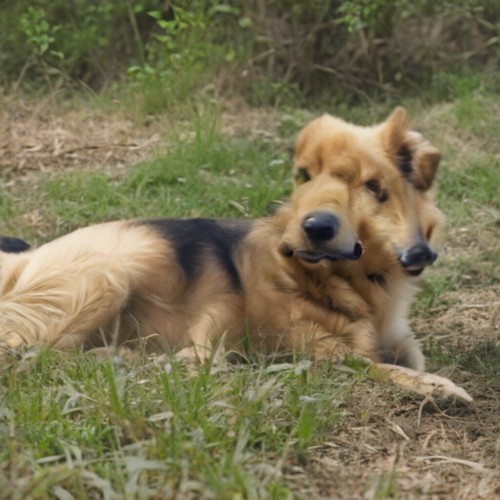} &
		\includegraphics[width=0.15\columnwidth]{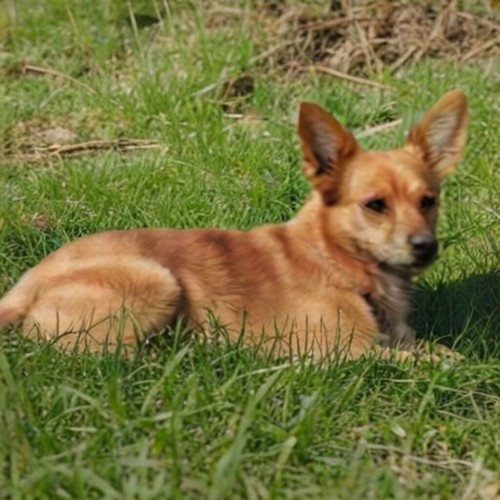} 
		\\
 
		 & {\footnotesize } & {\footnotesize 1} & {\footnotesize 2} & &{\footnotesize 3} & {\footnotesize 4} & {\footnotesize ours}  \\
		\rotatebox{90}{\tiny \phantom{} Cat $->$ Tiger} &
		\includegraphics[width=0.15\columnwidth]{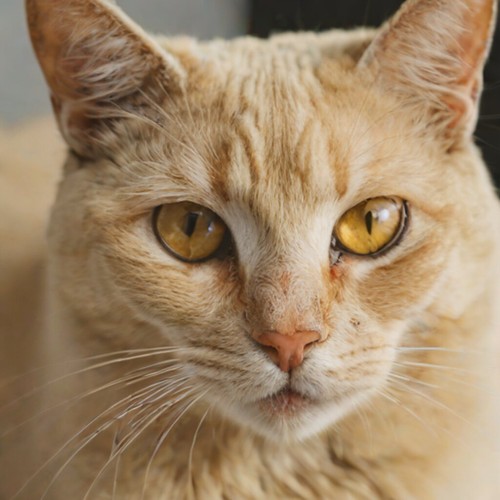} &
		\includegraphics[width=0.15\columnwidth]{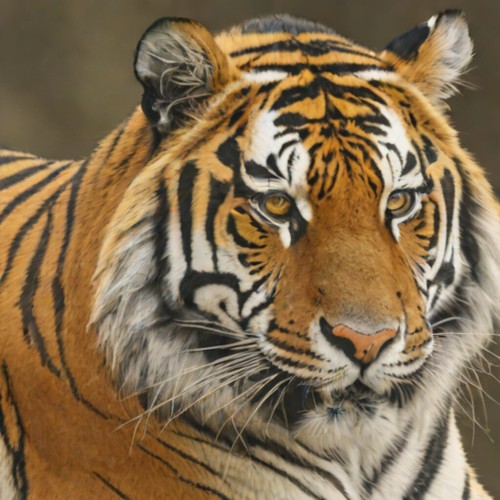} &
		\includegraphics[width=0.15\columnwidth]{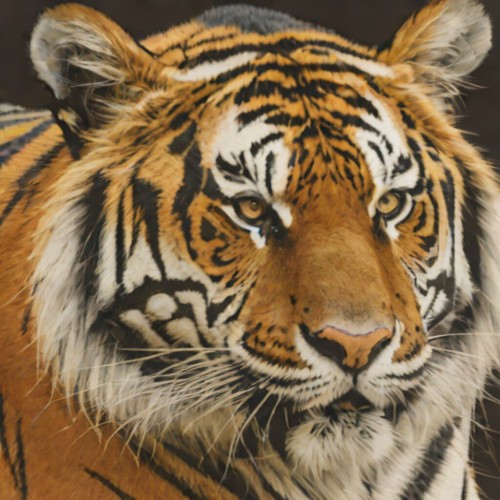} & &
		\includegraphics[width=0.15\columnwidth]{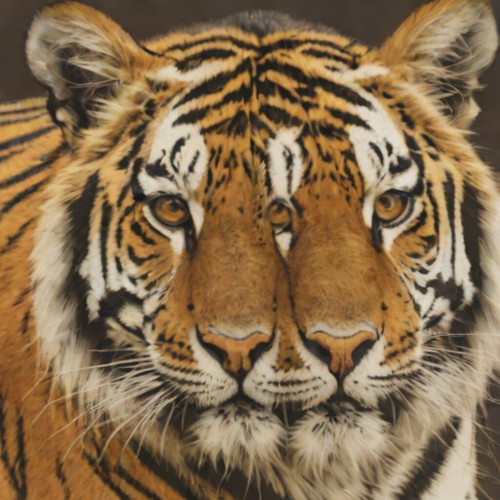} &
		\includegraphics[width=0.15\columnwidth]{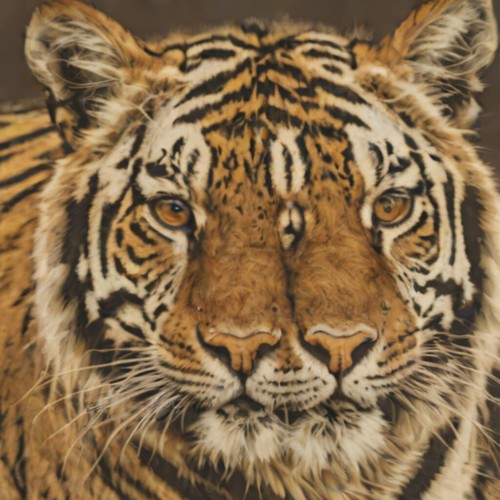} &
            \includegraphics[width=0.15\columnwidth]{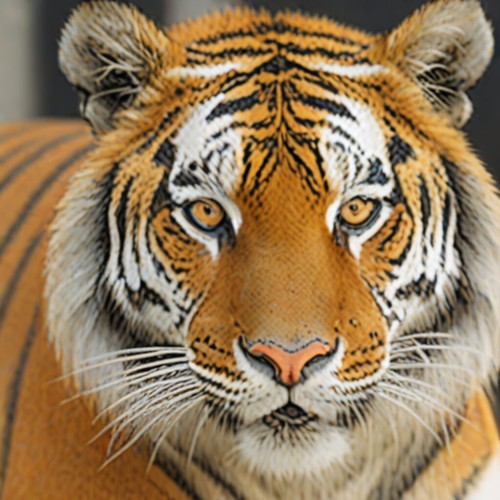}
		\\
  
	\end{tabular}
	\caption{ 
 Attention control methods \cite{hertz2022prompt,mokady2023null,parmar2023zero} exert an overly restrictive influence on the generation process, leading to insufficient changes in the image space (horse to unicorn) or the introduction of artifacts (fox to dog) in single-step diffusion model. Even if we only apply attention control on the initial generation steps of a four-step diffusion model, it leads to either inadequate preservation of structure (1 and 2) or the occurrence of artifacts (3 and 4), particularly in cases where the editing necessitates significant structural alterations.
    }
	\label{fig:attention}
\end{figure}

%% file: fig/long_short_prompt.tex
\begin{figure}[tb]
	\setlength{\tabcolsep}{1pt}	
        
	\begin{tabular}{ccccccc}
		 & {\scriptsize Original } & { \scriptsize Short Text } & { \scriptsize Detailed Text} \\
		\rotatebox{90}{\scriptsize \phantom{kkkkkkk} + Sunglasses} &
		\includegraphics[width=0.3\columnwidth]{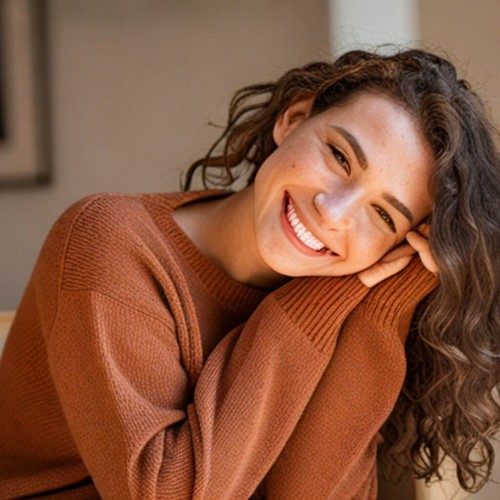} &
		\includegraphics[width=0.3\columnwidth]{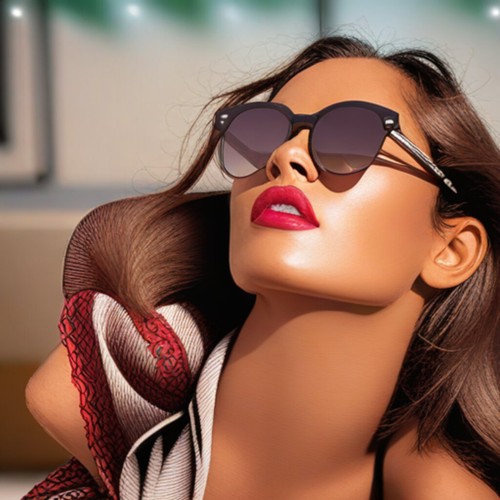} &
		\includegraphics[width=0.3\columnwidth]{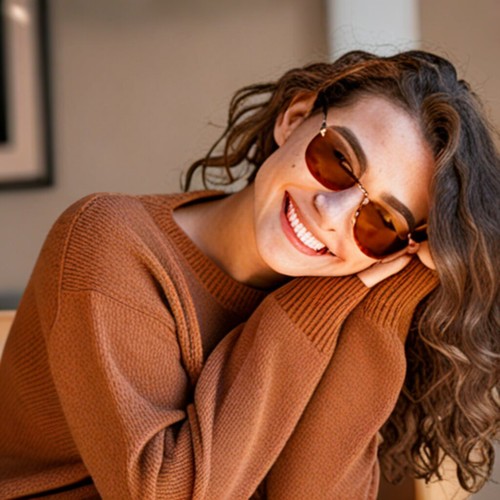} 
		\\
		\rotatebox{90}{\scriptsize \phantom{kkkkkkk} + Grey Color} &
		\includegraphics[width=0.3\columnwidth]{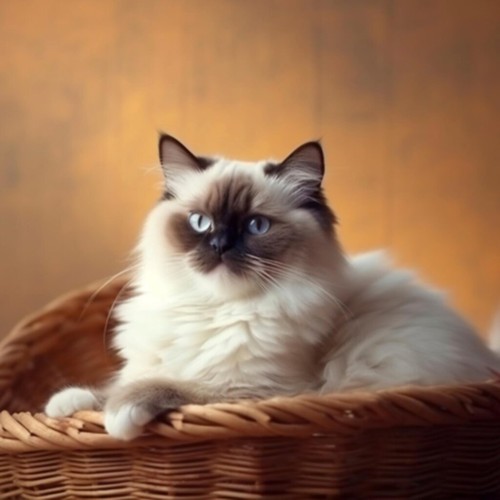} &
		\includegraphics[width=0.3\columnwidth]{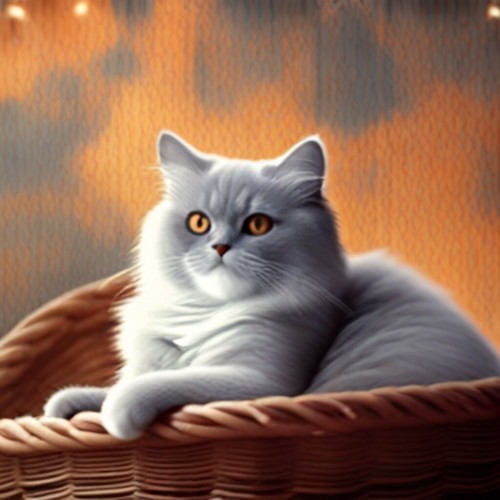} &
		\includegraphics[width=0.3\columnwidth]{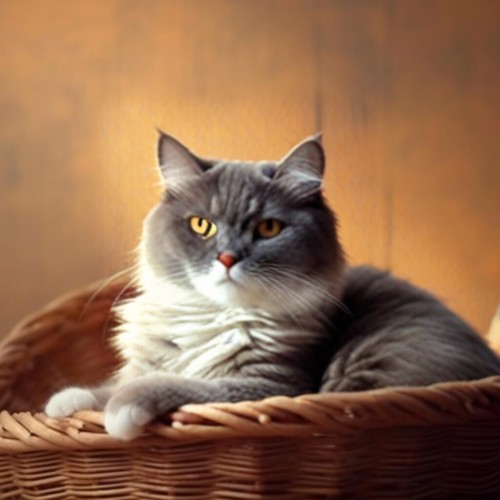} 
		\\
            
	\end{tabular}

        \raggedright
	\caption[Turbo]{ We demonstrate a detailed text prompt is necessary for real image disentangled editing. If we use short text prompt for both inversion and editing, the edited images will have large structure change (in sunglasses example) or artifacts in background regions (in grey color example). In contrast, if we use long detailed text prompt for inversion and editing, we can achieve disentangled edit without artifacts. }
    \scriptsize{\textcolor{blue}{Woman short text:} ``a woman \textcolor{red}{wearing sunglasses}.''} \\
   \scriptsize{\textcolor{blue}{Woman detailed text:} ``The image features a woman \textcolor{red}{wearing sunglasses} with curly hair, wearing a brown sweater and smiling. She is posing for the camera, with her arm resting on her head. The woman is the main focus of the scene, and her smile is the central element of the image. The sweater she is wearing is a warm, earth-toned color, and her curly hair''}\\
   \scriptsize{\textcolor{blue}{Cat short text:} ``a \textcolor{red}{grey color} cat.''} \\
   \scriptsize{\textcolor{blue}{Cat detailed text:} ``The image features a \textcolor{red}{grey color} cat sitting in a woven basket, which is placed on a wooden table. The cat appears to be looking at the camera, possibly posing for a picture. The basket is filled with hay, providing a comfortable and cozy spot for the cat to rest. The overall scene is warm and inviting, showcasing the cat's contentment''}

	\label{fig:long_short_prompt}
\end{figure}

%% file: fig/inversion_abl.tex
\begin{figure}[tb]
\centering
\includegraphics[width=1\columnwidth]{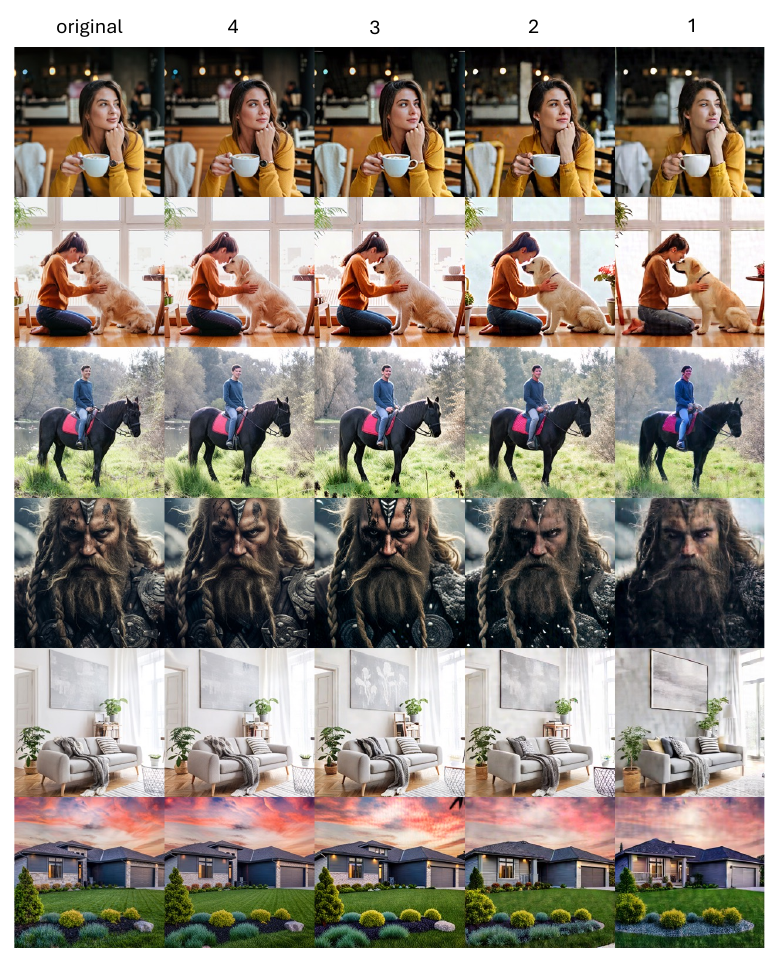} \\
\caption{Given the input image, we visualize the inversion results with different inversion steps. While single-step inversion captures much of the semantic information from the input image, it tends to inadequately preserve identity and local detail, and produces noticeable artifacts in the background region. In contrast, multi-step inversion significantly enhances the quality of the reconstruction. Specifically, a four-step inversion approach can achieve near-perfect reconstruction of the input image.}
	\label{fig:inversion_abl}
\end{figure}

%% file: fig/mask.tex
\begin{figure}[tb]
	\setlength{\tabcolsep}{1pt}	
	\begin{tabular}{ccccccc}
		 & {\scriptsize Original } & { \scriptsize Without Mask } & { \scriptsize With Mask} & { \scriptsize Attention Mask} \\
   
		\rotatebox{90}{\scriptsize \phantom{kk} Chicken $\rightarrow$ Salmon} &
		\includegraphics[width=0.25\columnwidth]{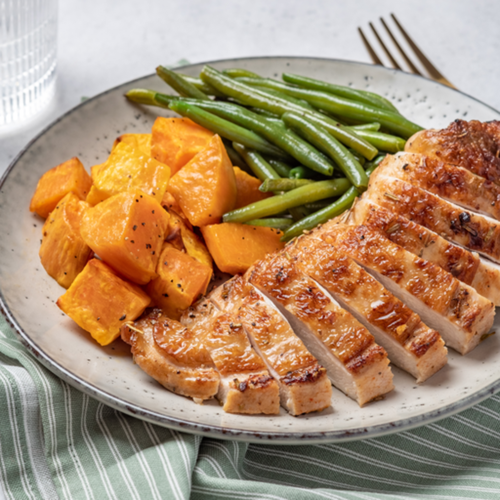} &
		\includegraphics[width=0.25\columnwidth]{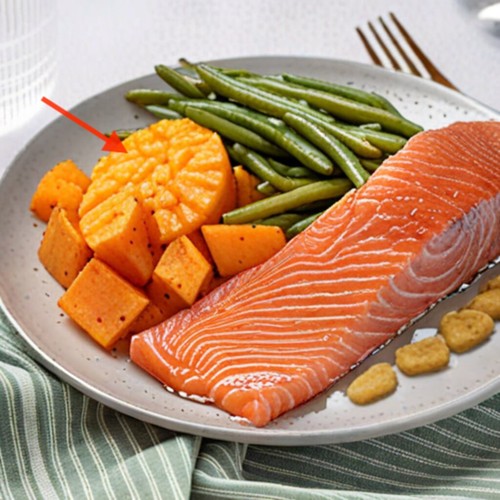}&
            \includegraphics[width=0.25\columnwidth]{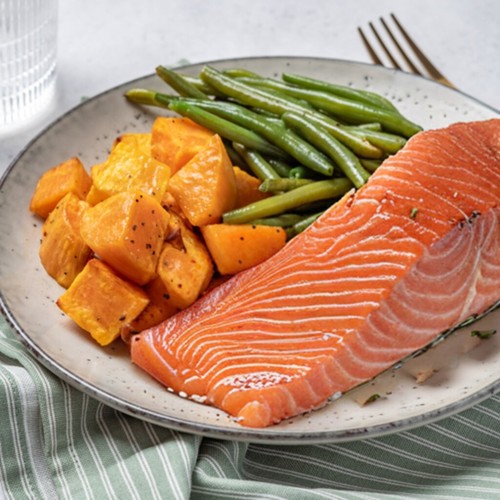} &
            \includegraphics[width=0.25\columnwidth]{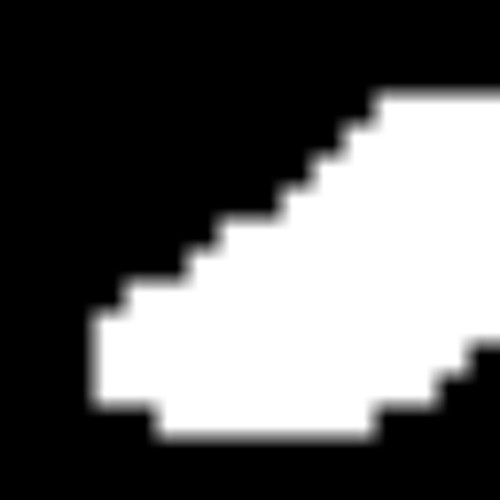} 
		\\

		\rotatebox{90}{\scriptsize \phantom{kkkk} Shirt $\rightarrow$ T-shirt} &
		\includegraphics[width=0.25\columnwidth]{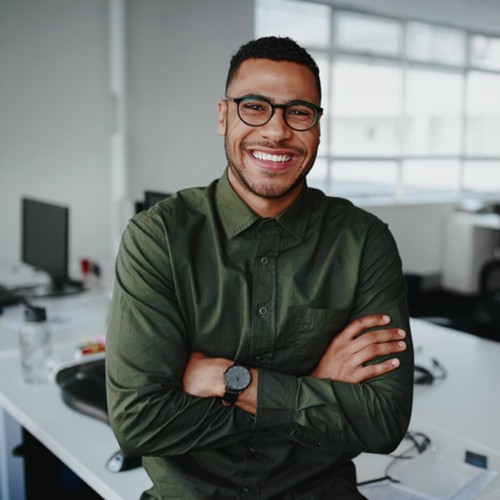} &
		\includegraphics[width=0.25\columnwidth]{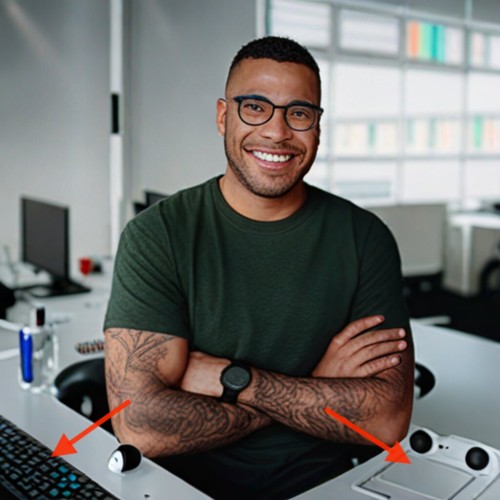} &
		\includegraphics[width=0.25\columnwidth]{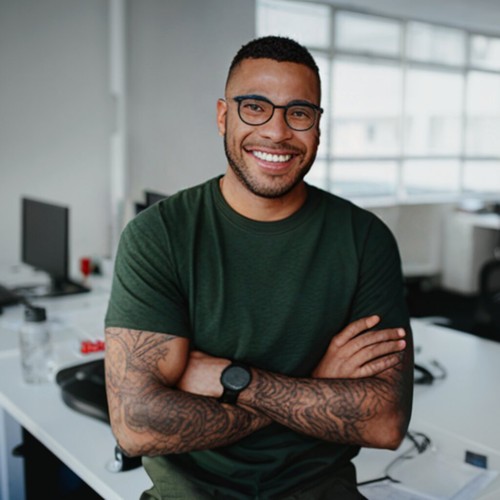} &
            \includegraphics[width=0.25\columnwidth]{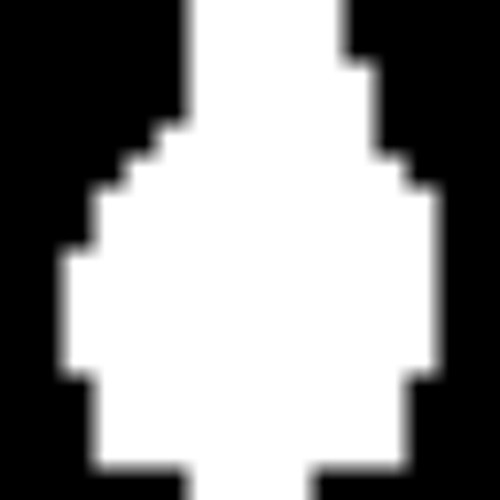} 
		\\
  
		\rotatebox{90}{\scriptsize \phantom{k} Shirt $\rightarrow$ Down Jacket} &
		\includegraphics[width=0.25\columnwidth]{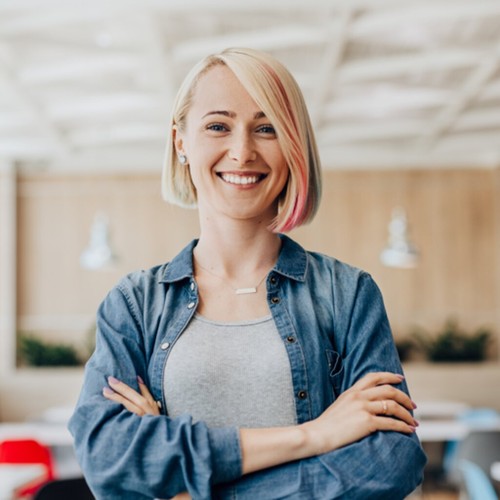} &
		\includegraphics[width=0.25\columnwidth]{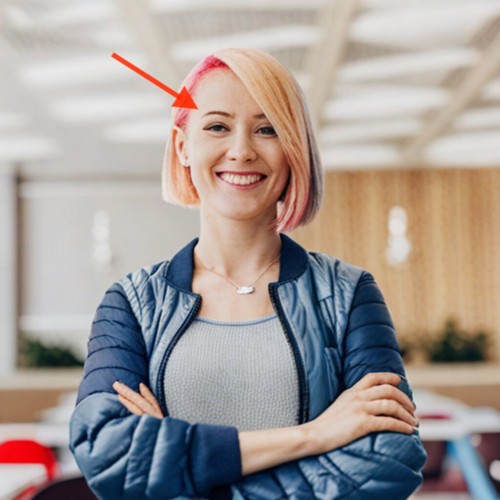} &
		\includegraphics[width=0.25\columnwidth]{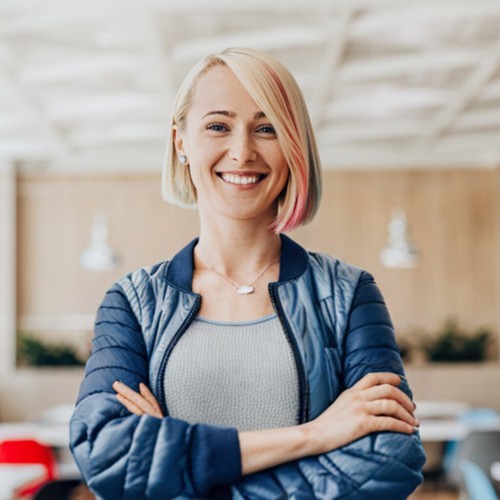} &
            \includegraphics[width=0.25\columnwidth]{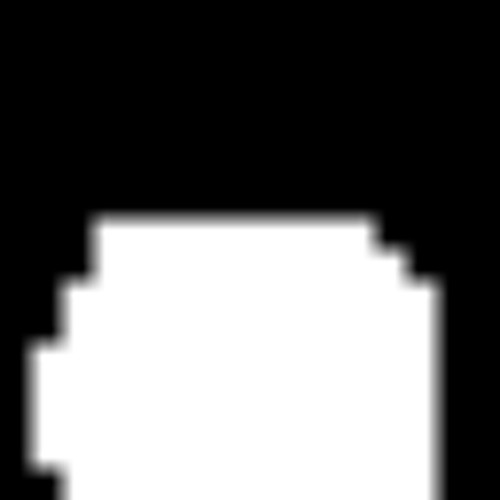} 
		\\

	\end{tabular}
        \raggedright
	\caption[Turbo]{
    In the absence of local masking, slight changes are observed in the background region after editing (indicating in red arrows). For instance, in the first example, the shape of sweet potatoes undergoes alteration, while in the second example, the shape of the background table change, and the identity of the face changes in the third example. Although attention mask is very rough, it significantly improves the background and identity preservation. It is worth to mention that manual mask will definitely help, since attention mask is too rough and sometime covers unedited region. For example, although we only modify the cloth region in the second example, the attention mask also covers the face region, which results in slightly identity changes. If users can provide a mask that only covers the cloth region, identity changes could be prevented.  
    }

	\label{fig:mask}
\end{figure}

%% file: fig/manual_mask.tex
\begin{figure}[tb]
	\setlength{\tabcolsep}{1pt}	
	\begin{tabular}{ccccccc}
		 & {\scriptsize Original } & { \tiny With Attention Mask } & { \scriptsize Attention Mask} & { \scriptsize With Manual Mask } & { \scriptsize Manual Mask }\\
   
		\rotatebox{90}{\scriptsize \phantom{k} Blond $\rightarrow$ Black} &
		\includegraphics[width=0.19\columnwidth]{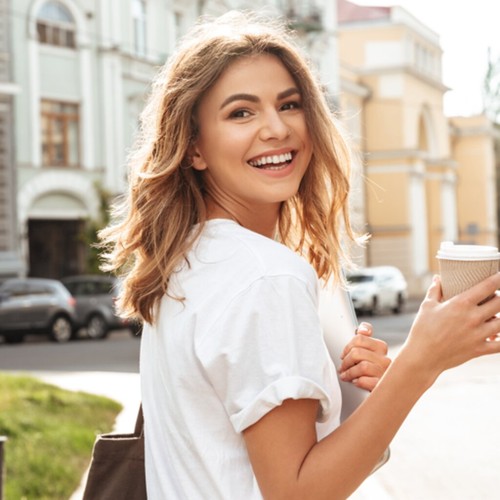} &
		\includegraphics[width=0.19\columnwidth]{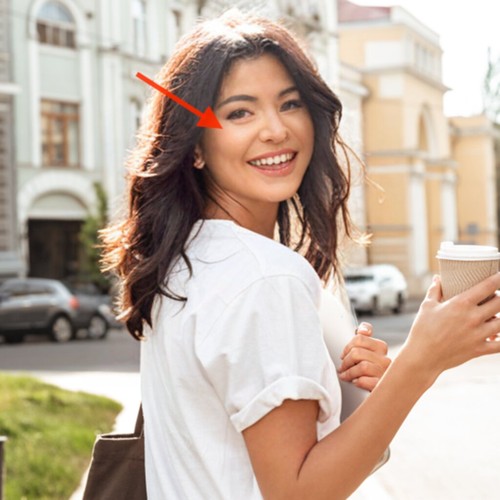}&
            \includegraphics[width=0.19\columnwidth]{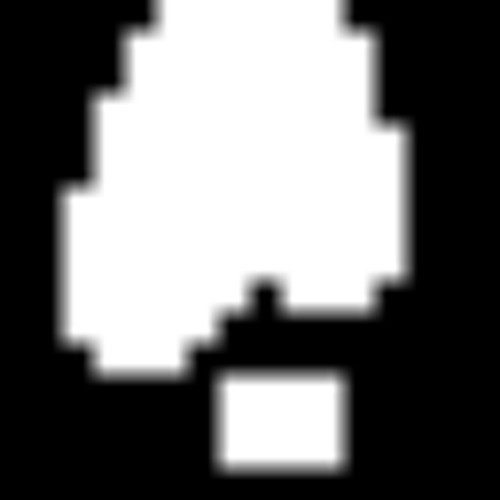} &
            \includegraphics[width=0.19\columnwidth]{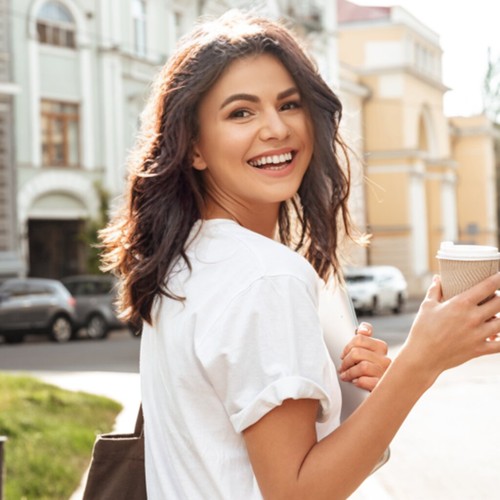} &
            \includegraphics[width=0.19\columnwidth]{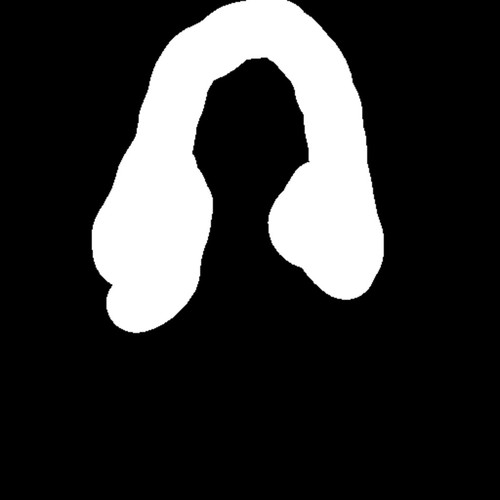}
		\\

		\rotatebox{90}{\scriptsize \phantom{k} Horse $\rightarrow$ Zebra} &
		\includegraphics[width=0.19\columnwidth]{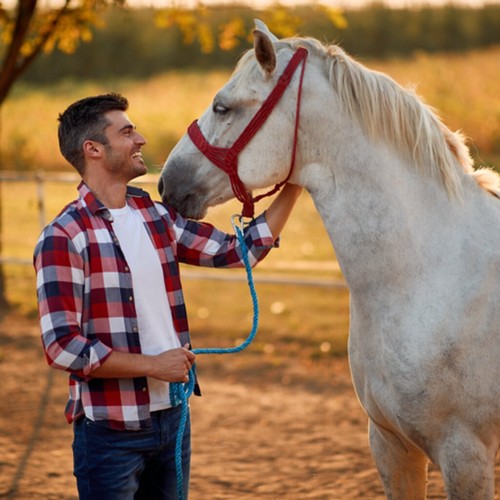} &
		\includegraphics[width=0.19\columnwidth]{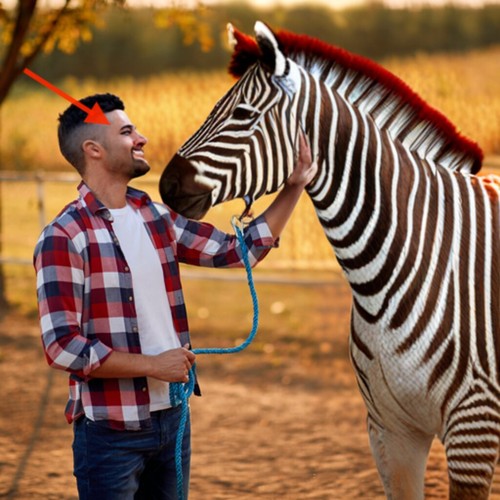}&
            \includegraphics[width=0.19\columnwidth]{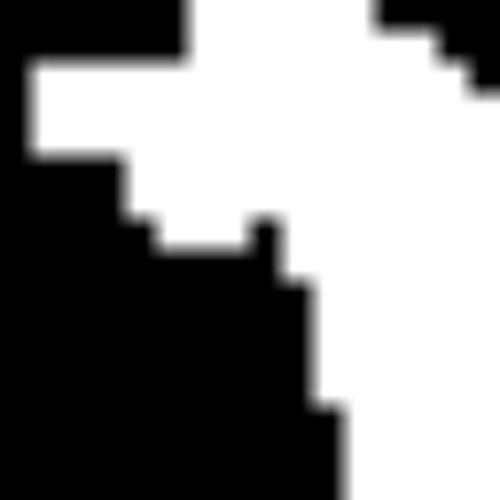} &
            \includegraphics[width=0.19\columnwidth]{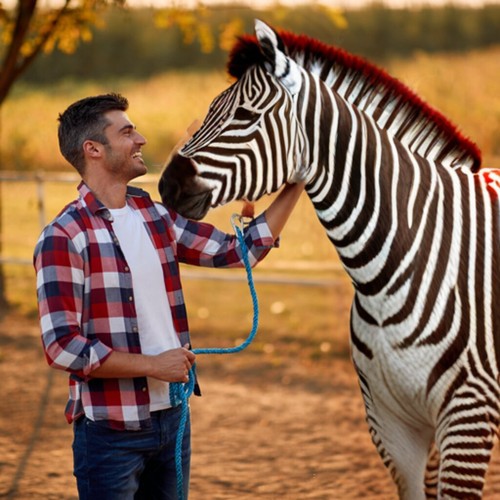} &
            \includegraphics[width=0.19\columnwidth]{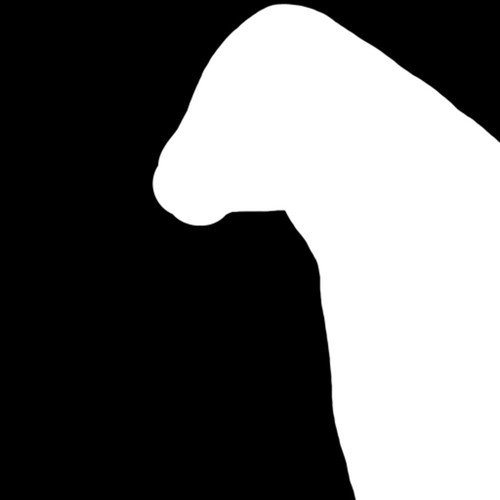}
		\\
  
	\end{tabular}
        \raggedright
	\caption[Turbo]{
While attention masks can generally identify the editing region, they often lack precision, leading to the inclusion of nearby regions. Consequently, nearby regions may undergo slight alterations post-editing, as indicated by the red arrows. Our approach enables users to upload a customized mask, which can be created manually or generated using an image segmentation model. We demonstrate that a manually created mask with a coarse outline can effectively minimize alterations in nearby pixels.}

	\label{fig:manual_mask}
\end{figure}

%% file: fig/renoise.tex
\begin{figure}[tb]
	\centering
	\setlength{\tabcolsep}{1pt}	
	\begin{tabular}{cccccccc}

		 {\scriptsize Original} &\phantom{\tiny k} &{\scriptsize ReNoise} & {\scriptsize Ours} &\phantom{\tiny k}&{\scriptsize ReNoise} & {\scriptsize Ours}  \\

		\includegraphics[width=0.18\columnwidth]{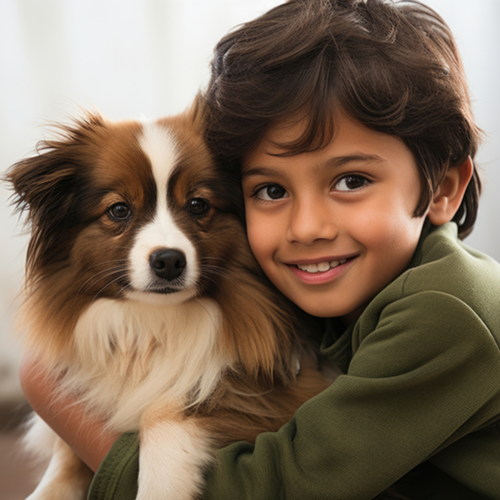} & &
              \includegraphics[width=0.18\columnwidth]{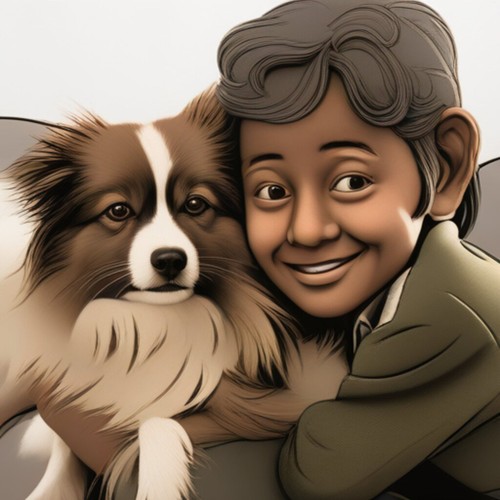} &
            \includegraphics[width=0.18\columnwidth]{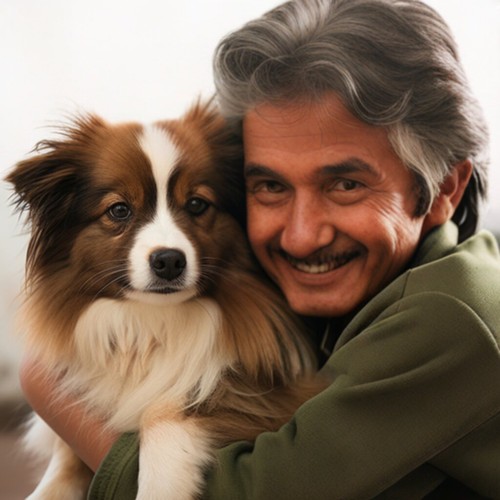} & &
            \includegraphics[width=0.18\columnwidth]{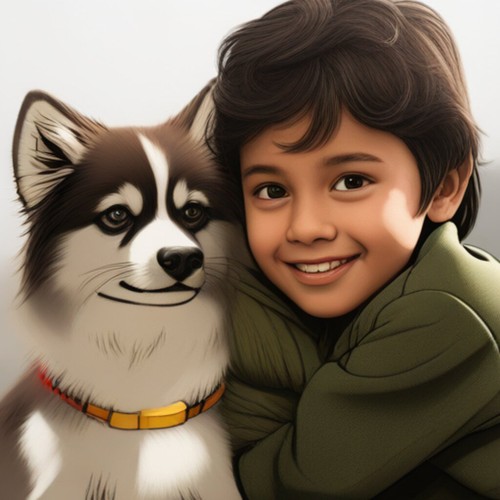} &
            \includegraphics[width=0.18\columnwidth]{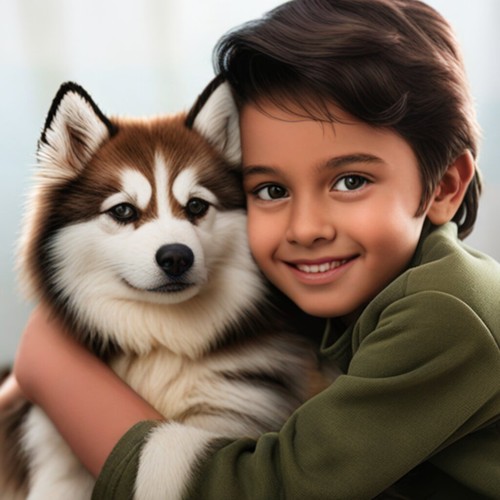} 
		\\
            & & \multicolumn{2}{c}{Young $\rightarrow$ Old} 
              & &\multicolumn{2}{c}{Dog $\rightarrow$ Husky} \\

		\includegraphics[width=0.18\columnwidth]{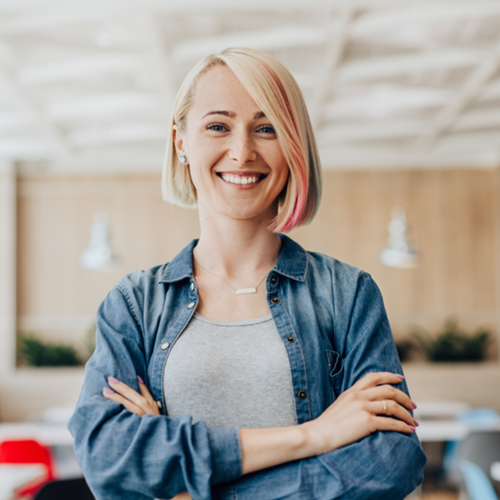} & &
            \includegraphics[width=0.18\columnwidth]{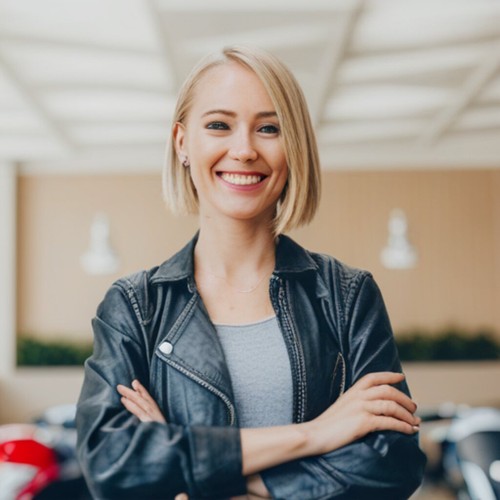} &
            \includegraphics[width=0.18\columnwidth]{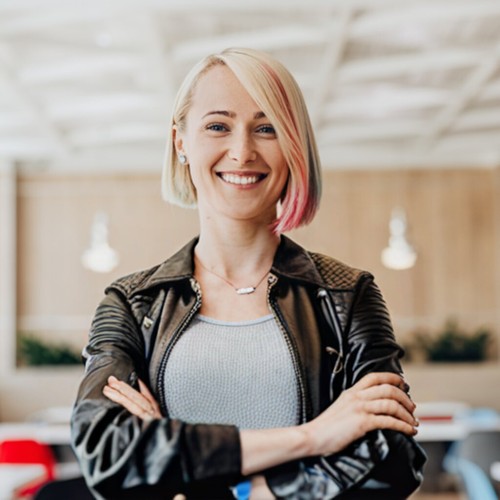} &&
            \includegraphics[width=0.18\columnwidth]{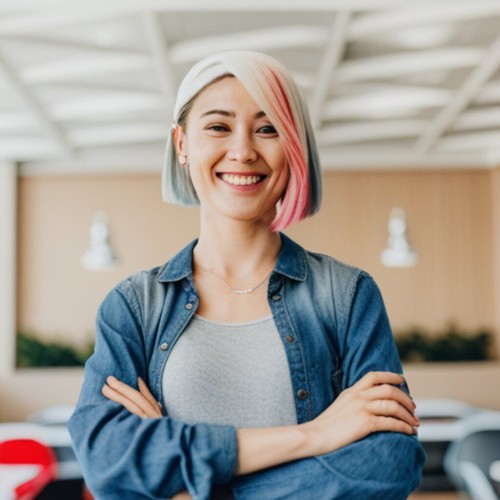} &
            \includegraphics[width=0.18\columnwidth]{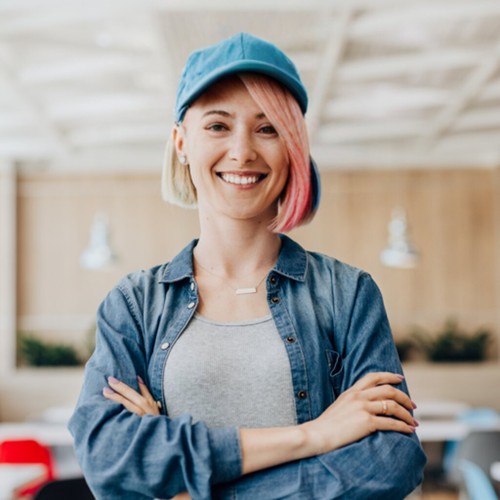} \\
            & & \multicolumn{2}{c}{Shirt $\rightarrow$ Leather Jacket} 
              & &\multicolumn{2}{c}{+Hat} \\

		\includegraphics[width=0.18\columnwidth]{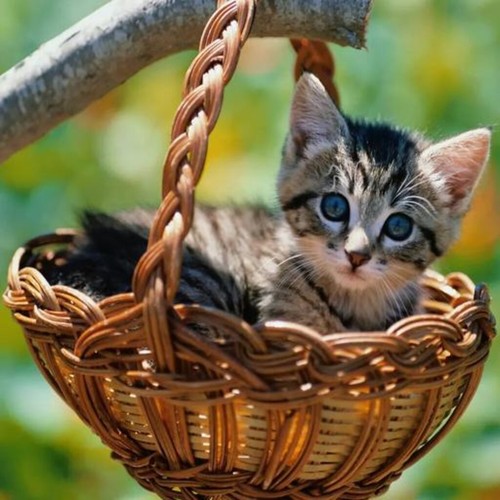} &&
            \includegraphics[width=0.18\columnwidth]{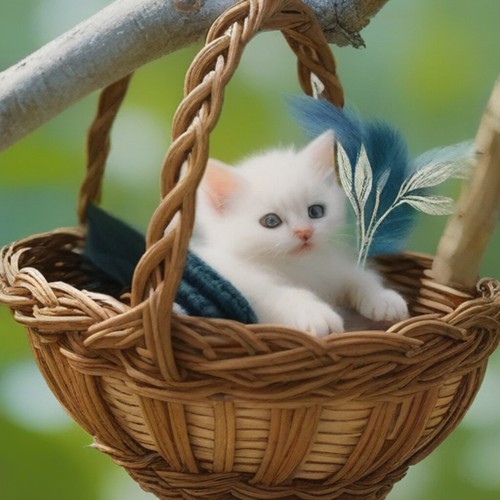} &
            \includegraphics[width=0.18\columnwidth]{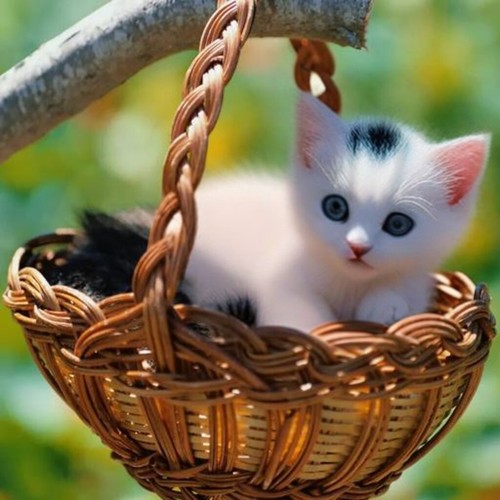} & &
            \includegraphics[width=0.18\columnwidth]{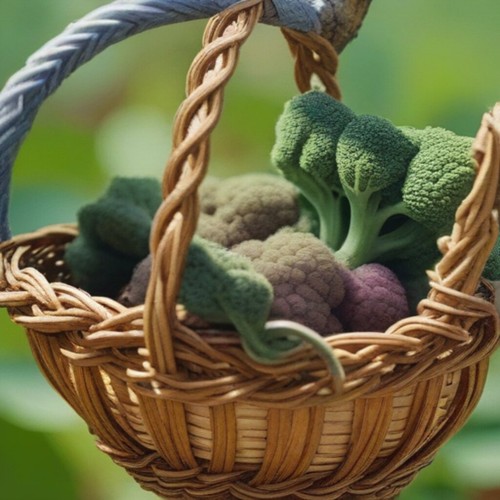} &
            \includegraphics[width=0.18\columnwidth]{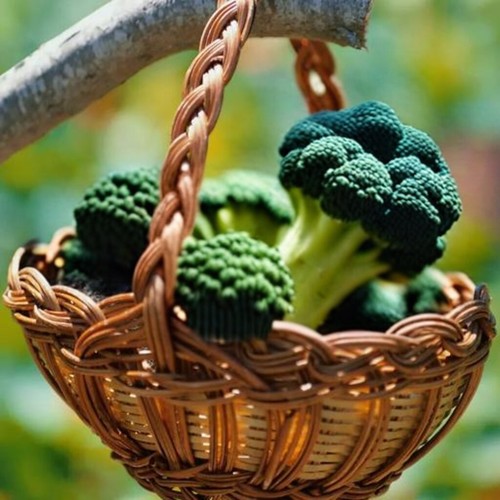} 
		\\
            & & \multicolumn{2}{c}{ Black $\rightarrow$ White} 
              & &\multicolumn{2}{c}{ Kitten $\rightarrow$ Broccoli} \\
              
		\includegraphics[width=0.18\columnwidth]{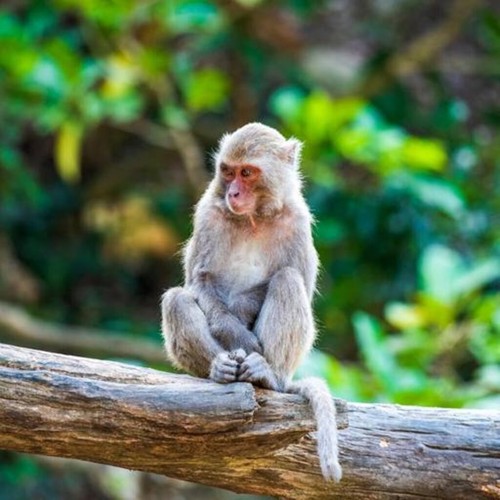} &&
            \includegraphics[width=0.18\columnwidth]{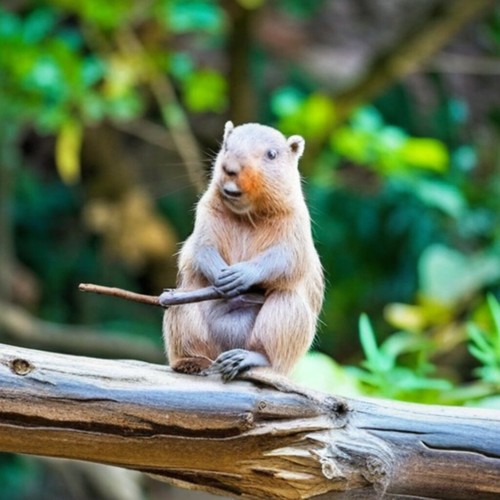} &
            \includegraphics[width=0.18\columnwidth]{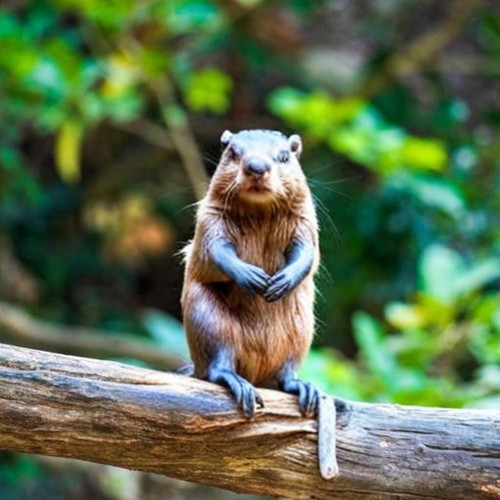} &&
            \includegraphics[width=0.18\columnwidth]{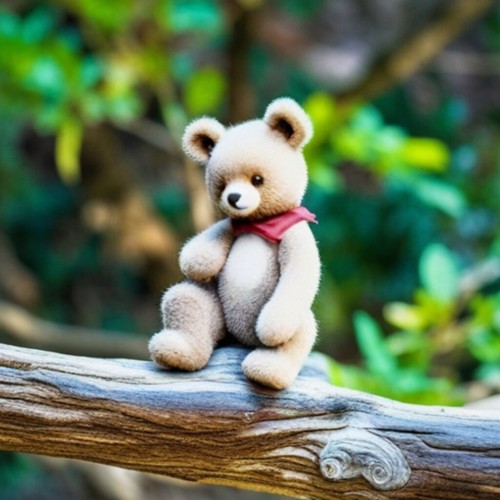} &
            \includegraphics[width=0.18\columnwidth]{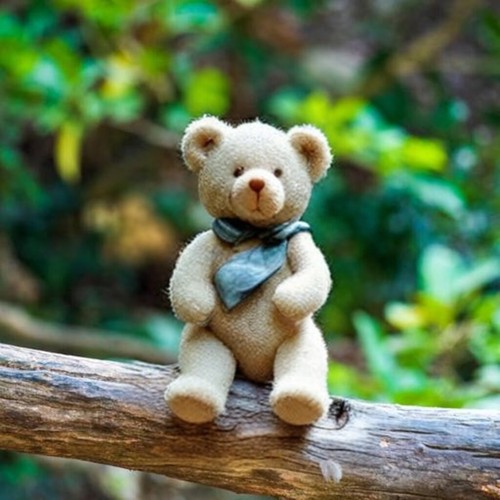} 
		\\
            & & \multicolumn{2}{c}{Monkey $\rightarrow$ Beaver} 
              & &\multicolumn{2}{c}{ Monkey $\rightarrow$ Teddy Bear} \\

	\end{tabular}
	\caption{
        Compare to ReNoise method, our method generates realistic edited images without artifacts (dog $\rightarrow$ husky, young $\rightarrow$ old), maintains the face identity better (shirt $\rightarrow$ leather jacket), can perform large structure change (+hat). For the edits that ReNoise works well (we take the kitten and monkey images from ReNoise demo), our method generates comparable results.}
	\label{fig:renoise}
\end{figure}

%% file: fig/failure.tex
\begin{figure}
	\centering
	\setlength{\tabcolsep}{1pt}	
	\begin{tabular}{cccccc}
  		   & \tiny Run $\rightarrow$ Sit & & \tiny + Blooming  \\
		\includegraphics[width=0.23\columnwidth]{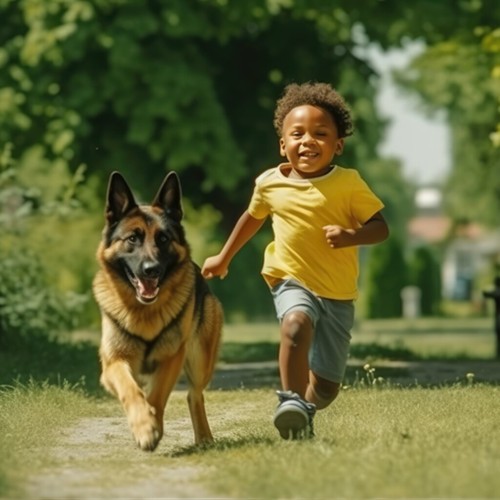} &
            \includegraphics[width=0.23\columnwidth]{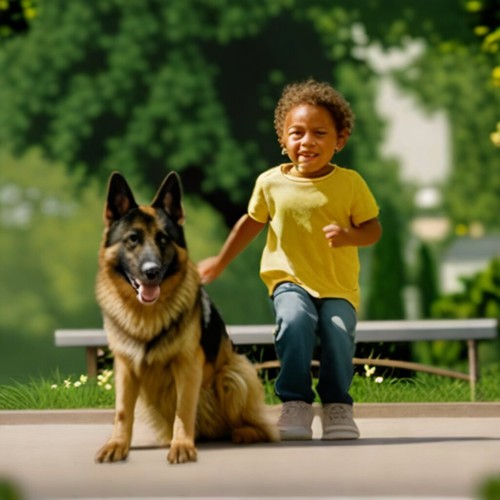} &
		\includegraphics[width=0.23\columnwidth]{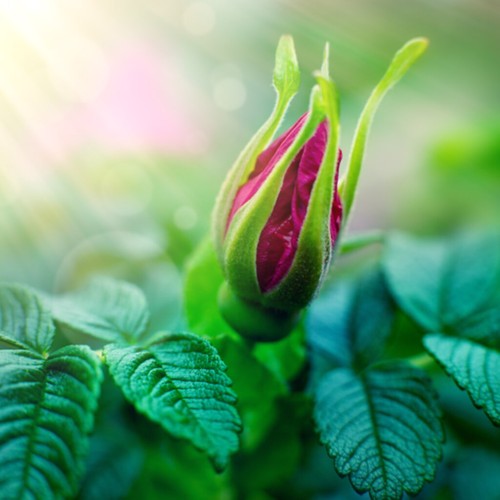} &
            \includegraphics[width=0.23\columnwidth]{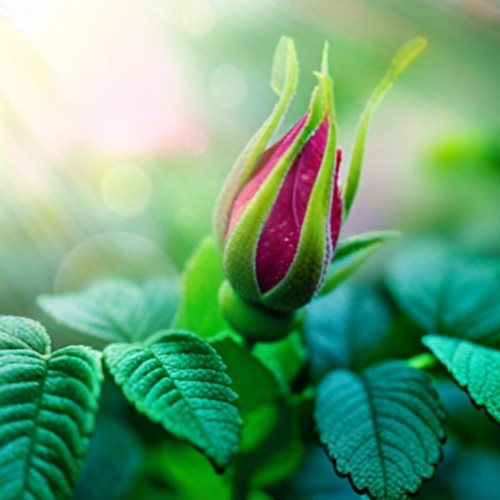} 
		\\
	\end{tabular}
        \vspace{-2mm}
	\caption{ Failure cases. Our method struggles with the difficult cases of large pose changes (e.g. run to sit, bud to blooming)}
        \vspace{-2mm}
	\label{fig:failure}
\end{figure}